%% file: iclr2027_conference.tex
\documentclass{article} 
\usepackage{iclr2027_conference,times}

\input{math_commands.tex}

\usepackage{wrapfig}
\usepackage{caption}

\usepackage[hidelinks]{hyperref}
\usepackage{url}
\usepackage{amssymb} 
\usepackage{graphicx} 

\usepackage{multirow}
\usepackage{makecell}

\usepackage{algorithm}
\usepackage{algpseudocode}

\usepackage{amsmath}
\usepackage{booktabs}
\usepackage{multirow}
\usepackage{threeparttable}
\usepackage{pifont}
\usepackage{tabularx}
\usepackage{xcolor,colortbl}
\usepackage{capt-of}
\definecolor{tabgray}{gray}{0.95}
\providecommand{\MainSource}[1]{\mbox{\citealt{#1}}}

\title{OpenNWM: A Generalist Navigation World Model with Latent Action Pretraining}

\author{%
\begin{minipage}[t]{\dimexpr\textwidth-2\tabcolsep\relax}
\raggedright\normalfont\bfseries
Dujun Nie\textsuperscript{1,2*}, Ruiqi Song\textsuperscript{1,5*}, Zixuan Wang\textsuperscript{1,2},
Hanqing Wang \textsuperscript{3},
Yifan Han\textsuperscript{1,2}, \\[3pt] Baiyong Ding\textsuperscript{1},  Wenzhao Zheng\textsuperscript{4},  Hangbin Wu\textsuperscript{5}, Long Chen\textsuperscript{1,6\textdagger}
\normalfont
\\[6pt]
\textsuperscript{1} CASIA \quad \textsuperscript{2} UCAS \quad \textsuperscript{3} HKUST(GZ) \quad \textsuperscript{4} UC Berkeley \quad
\textsuperscript{5} Tongji University \quad \textsuperscript{6} Waytous
\\
{\small \textsuperscript{*}Equal Contribution \quad \textsuperscript{\textdagger}Corresponding author }\\[3pt]
{\small \textbf{Project Page:} \url{https://openspace-lab.github.io/OpenNWM/}}
\end{minipage}%
}

\iclrfinalcopy 
\begin{document}

\maketitle

\begin{abstract}
Navigation world models predict future observations conditioned on navigation actions to support planning, but their generalization across real-world environments remains constrained by the limited coverage of action-labeled data. Action-free navigation videos provide diverse visual experience, yet lack the recorded actions required for action-conditioned training. We introduce OpenNWM, a generalist navigation world model with latent action pretraining. It learns universal latent actions from large-scale action-free videos through visual prediction pretraining, and further aligns them with physical motions via action-conditioned post-training. To support video pretraining, we curate NavAnywhere, a large-scale action-free navigation video dataset comprising 17.5 million visual observations from 15 heterogeneous sources. It captures navigation by ground robots, humans, and drones, with diverse viewpoints and motion patterns across five broad scene types: residential areas, public and commercial spaces, urban settings, parks, and natural landscapes. Experiments demonstrate improved visual prediction on in-domain and out-of-domain benchmarks, together with improved navigation performance. These results show that latent-action pretraining enables diverse, action-free videos to improve dynamics learning and generalization for open-world navigation.
\end{abstract}

\input{main/Introduction}
\input{main/Related}
\input{main/Method}

\input{main/Experiments}

\section*{AI Use Statement}
In this work, we have not used generative AI tools for any tasks with required disclosure, and all such tasks are not applicable to this work. We used generative AI tools only to polish the writing of parts of the manuscript. The authors reviewed all AI-assisted edits to ensure that they preserve the intended technical meaning and contain no unsupported claims. We take responsibility for the final content of this work, including text, claims, or artifacts produced with the aid of generative AI.

\section*{Reproducibility Statement}
For reproducibility, the main paper and appendices detail our methodology and implementation. We will release the complete source code and model weights.



\bibliography{iclr2027_conference}
\bibliographystyle{iclr2027_conference}

\newpage
\appendix
\input{main/Appedix}

\end{document}

%% file: math_commands.tex
\usepackage{amsmath,amsfonts,bm}

\def\eqref#1{equation~\ref{#1}}

\def\1{\bm{1}}

\DeclareMathAlphabet{\mathsfit}{\encodingdefault}{\sfdefault}{m}{sl}
\SetMathAlphabet{\mathsfit}{bold}{\encodingdefault}{\sfdefault}{bx}{n}



%% file: main/Introduction.tex
\begin{figure*}[h]
    \centering
    \vspace{-6pt}
    \includegraphics[width=0.9\textwidth]{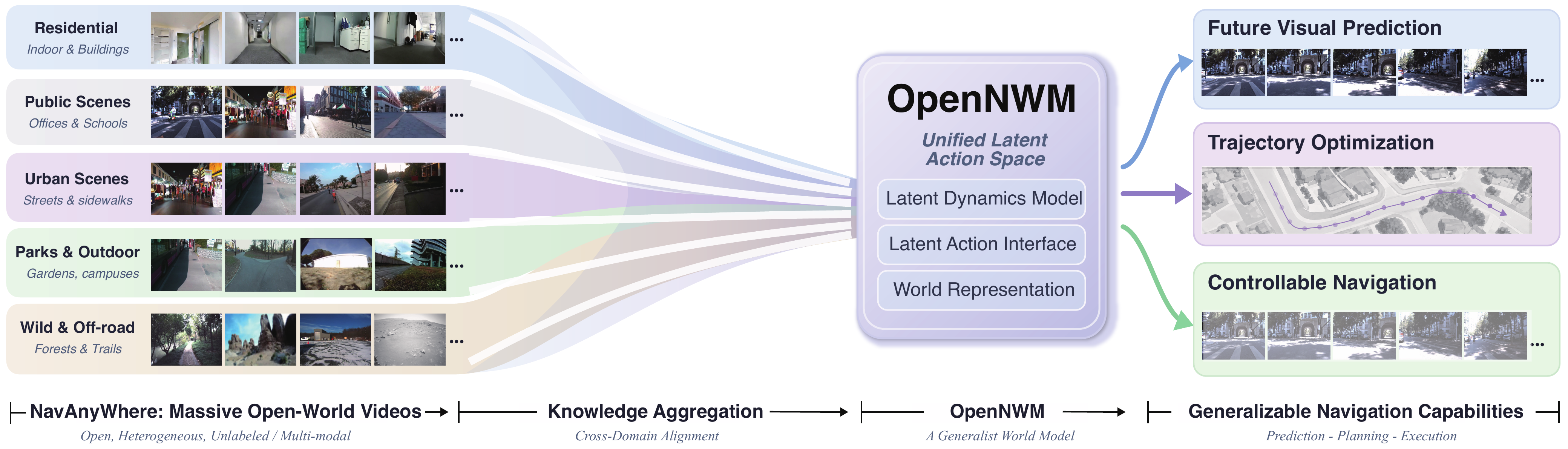}
    \caption{\textbf{Overview of OpenNWM.}
OpenNWM learns navigation dynamics from heterogeneous action-free videos via latent-action pretraining, then adapts to real waypoint actions for action-conditioned prediction and planning.}
    \vspace{-12pt}
    \label{fig:overview}
\end{figure*}

\section{Introduction}
\label{sec:introduction}

Visual navigation is a fundamental capability that enables embodied agents to autonomously navigate and accomplish tasks in real world. Recent navigation policies generate actions from visual observations conditioned on goal images~\citep{shah2023gnm, shah2023vint, sridhar2024nomad} or language instructions~\citep{spikevla}, demonstrating promising generalization across diverse environments and embodiments. Beyond direct policy learning, navigation world models explicitly model environmental dynamics by predicting future visual observations conditioned on candidate actions~\citep{zhu2026astra, yu2025egosim}, enabling agents to imagine and evaluate potential trajectories for flexible planning~\citep{bar2025navigation, zhang2026rae}. WMNav~\citep{wmnav} integrates vision-language models into a
world-model framework, using semantic predictions and memory feedback
to guide object-goal navigation. However, existing NWMs rely heavily on action-labeled navigation trajectories, limiting their ability to exploit the breadth of visual experience required for open-world navigation.

Action-labeled trajectories provide a direct connection between navigation actions and their visual consequences, but collecting them across diverse environments and embodiments is costly~\citep{huang2025vid2world}. In contrast, large-scale action-free videos cover diverse egocentric, driving, and scene exploration scenarios without requiring recorded actions~\citep{grauman2022ego4d, liu2025citywalker, ling2024dl3dv, akhtyamov2025egowalk}. These videos contain rich visual and temporal information, yet do not directly supervise the relationship between executable actions and observed transitions~\citep{wan2026learning}. NWM partially addresses this gap through time-conditioned prediction on unlabeled videos~\citep{bar2025navigation}, but a temporal offset specifies \emph{when} to predict rather than explicitly describing the motion underlying a transition. This reveals a central tension: action-labeled trajectories provide control supervision but limited coverage, while action-free videos provide broad visual experience without explicit control signals.

Latent action models (LAMs) provide a promising direction by inferring proxy actions from visual transitions, enabling learning from videos without explicit action annotations~\citep{schmidt2024lapo, ye2025lapa, gao2025adaworld, garrido2026learning}. However, reconstruction-based latent actions may capture appearance changes and external dynamics rather than agent-controllable motion~\citep{nikulin2025latent}. Recent approaches such as villa-X and LatBot incorporate physical supervision to better connect visual transitions with executable motion~\citep{chen2026villax, li2026latbot}. For navigation world models, this raises two connected questions: \emph{what supervision makes latent actions useful for learning navigation dynamics, and how can the resulting predictive dynamics be adapted to real navigation actions?} Addressing both questions is important for turning diverse video experience into effective action-conditioned prediction and planning.

To this end, we introduce OpenNWM (see Figure~\ref{fig:overview}), a generalist navigation world model built upon a universal latent-action space for open-world navigation. OpenNWM learns Universal Navigation Latent Actions (UNLAs) that work as an intermediate training signal for learning predictive dynamics from videos. By optimizing a unified visual-physical objective, OpenNWM learns latent representations that support both future visual prediction and physical motion prediction.

We construct {NavAnywhere}, an action-free navigation video dataset comprising {17.5 million visual observations from 15 sources}. It aggregates public datasets and online videos spanning ground robots, humans, and drones. The dataset covers residential environments, public and commercial spaces, urban and transportation settings, parks and gardens, and natural and off-road environments. 
We study the effects of latent-action pretraining and LAM supervision on navigation world modeling. As prediction quality alone does not ensure effective decision-making~\citep{zhang2026world}, we evaluate OpenNWM on in-domain and out-of-domain prediction, autoregressive rollouts, and planning under real actions.
Our main contributions are as follows:
\begin{itemize}
    \item We introduce OpenNWM, a generalist navigation world model that learns a universal latent-action space from large-scale heterogeneous visual experiences, enabling scalable navigation learning across diverse environments, embodiments, and motion patterns.
    \item We curate NavAnywhere, a large-scale heterogeneous navigation dataset comprising 17.5 million visual observations from 15 sources across diverse environments and embodiments. We further propose Universal Navigation Latent Actions (UNLAs), which disentangle agent motion from environmental dynamics and unify heterogeneous visual experiences with navigation signals within a shared latent space.
    \item OpenNWM achieves state-of-the-art perceptual prediction and downstream navigation performance on the evaluated benchmarks. Systematic ablations further characterize how latent-action pretraining and different LAM supervision strategies contribute to dynamics modeling and cross-domain generalization.
\end{itemize}

%% file: main/Related.tex
\section{Related Work}
\subsection{From Navigation Policies to Navigation World Models}
\label{sec:related_navigation_models}
Learning-based visual navigation has evolved from task-specific controllers to generalist policies that transfer across environments and embodiments. Early visual-goal navigation systems combined learned reachability models with topological memory, enabling long-horizon navigation from offline visual experience~\citep{savinov2018sptm,shah2021ving,shah2021recon,shah2022viking}. This line of work subsequently converged on shared navigation interfaces: GNM provided a unified interface across heterogeneous robot trajectories through embodiment-agnostic waypoint prediction~\citep{shah2023gnm}; ViNT extended this interface with large-scale Transformer pretraining~\citep{shah2023vint}; and NoMaD employed diffusion-based action generation to unify goal-conditioned navigation and goal-agnostic exploration~\citep{sridhar2024nomad}. Together, these advances highlighted cross-dataset training for transferable navigation policies.

In parallel, pretrained vision--language models extend navigation beyond visual goals to semantic instructions and general tasks, integrating language, vision, and navigation modules for long-horizon reasoning and execution~\citep{shah2022lmnav,zhou2023navgpt}. Recent video-based navigation models integrate temporal visual context, language understanding, and action prediction, improving generalization across tasks, datasets, and sim-to-real transfer~\citep{zhang2024navid,zhang2024uninavid}. NaVILA further couples high-level VLA reasoning with embodiment-specific locomotion skills, enabling language-guided navigation for legged robots~\citep{cheng2025navila}. Meanwhile, automatic extraction of motion supervision from web-scale walking and driving videos expands the coverage of scenes and behaviors available for policy learning~\citep{liu2025citywalker}.

Unlike policy-centric methods that implicitly encode dynamics, Navigation World Models explicitly predict action-conditioned futures for planning~\citep{bar2025navigation}. OpenNWM adopts latent actions as a shared control interface, enabling the learning of controllable dynamics jointly from action-labeled trajectories and action-free videos.

\subsection{Unified Embodied Learning from Heterogeneous Data}
\label{sec:related_heterogeneous_data}
The development of generalist embodied models increasingly relies on jointly learning from diverse experiences across tasks, environments, and robot embodiments. Studies on large-scale robot datasets demonstrate that cross-dataset training facilitates knowledge transfer across robotic platforms, while flexible architectures accommodate heterogeneous observation modalities and action spaces~\citep{oneill2024openx,octo2024octo}. However, such unified learning paradigms still assume that every trajectory is paired with executable robot actions. Their scalability is therefore constrained by the high costs of robot deployment, teleoperated data collection, and alignment across heterogeneous action spaces.

Human egocentric recordings and Internet videos provide large-scale embodied experiences, spanning substantially more diverse scenes and behaviors. Recent studies show that carefully curated human videos can yield more transferable pretrained representations than an equivalent amount of robot data~\citep{ma2026humanscale}. Large-scale human-video pretraining also enables world models to acquire broad interaction and environmental priors before adaptation to specific robotic platforms~\citep{dreamdojo}. However, human and Internet videos lack robot-compatible actions, while embodiment-specific pose estimation and motion retargeting hinder a unified control interface.

Latent actions unify heterogeneous embodied data beyond predefined action spaces, supporting action-free pretraining, task transfer, and visual-action alignment~\citep{schmidt2024lapo,ye2025lapa,bu2025univla,lin2026pixels,chen2026villax,li2026latbot}. OpenNWM extends latent-action learning beyond manipulation and task-specific models to generalist navigation, unifying action-free videos and heterogeneous motion signals through a controllable world-model interface.

%% file: main/Method.tex
\section{Method}
\label{sec:method}
\begin{figure*}[t]
    \centering
\includegraphics[width=0.95\textwidth]{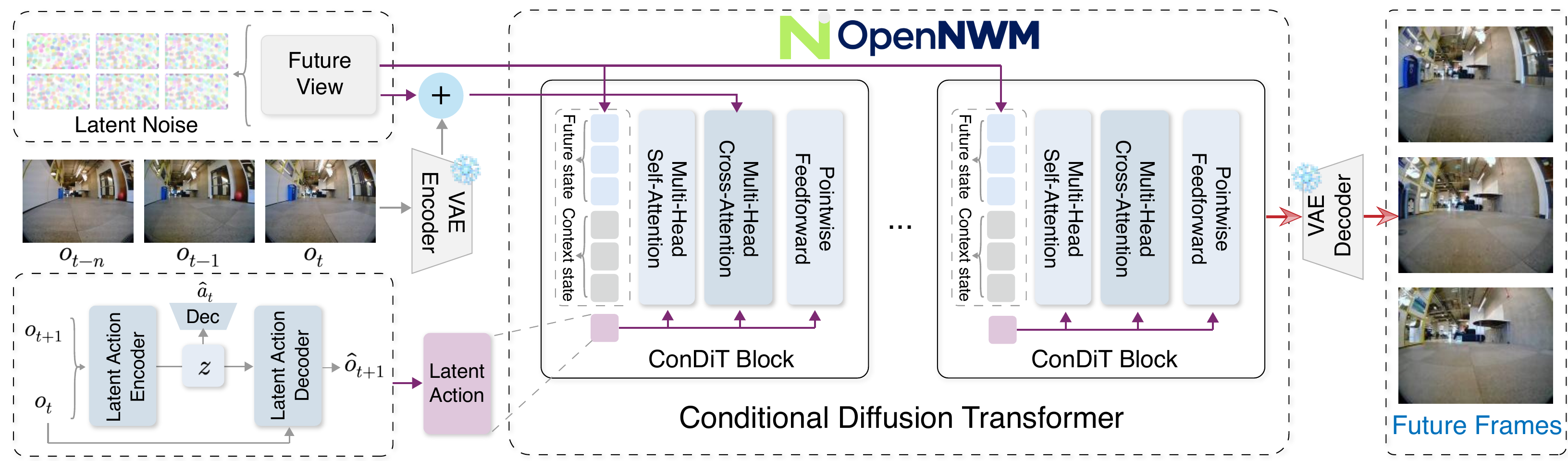}
\caption{
\textbf{Framework of OpenNWM.}}
    \label{fig:framework}
     \vspace{-12pt}
\end{figure*}

\subsection{Problem Setup}
\label{sec:formulation}
\textbf{Dataset.}~
Given $N$ heterogeneous visual observations, we construct
$\mathcal{D}=\{(\mathbf{x}_i,\mathbf{y}_i,c_i)\}_{i=1}^{N}$,
where $\mathbf{x}_i=(x_{i,0},\ldots,x_{i,T_i})$ denotes a visual sequence,
$\mathbf{y}_i$ contains optional camera poses, and $c_i$ encodes the data source, embodiment,
and sampling rate. For action-free videos, $\mathbf{y}_i=\varnothing$.

\textbf{Learning Target.}~
Given visual observations up to time $\tau$, we denote the encoded
visual states of the most recent $h$ frames as
$\mathbf{s}_{\tau-h+1:\tau}$, where $h$ is the visual context length. For a prediction horizon $k$, the target
time is defined as $m=\tau+k$. 
A latent action is inferred from the visual transition as
$z_{\tau\rightarrow m}\sim q_{\phi}(\cdot\mid x_{\tau},x_m)$,
where $q_{\phi}$ discovers a compact latent representation of visual changes
without requiring explicit action annotations.
Conditioned on the inferred latent action, OpenNWM learns future visual
dynamics by optimizing:
\begin{equation}
\label{eq:opennwm_target}
\mathcal{J}_{\mathrm{pred}}(\theta,\phi)
=
\mathbb{E}_{z_{\tau:m}\sim
q_{\phi}(\cdot\mid x_{\tau},x_m)}
\left[
\log
p_{\theta}
(s_m\mid
\mathbf{s}_{\tau-h+1:\tau},
z_{\tau:m},
k)
\right]
\end{equation}

Here, $p_{\theta}$ denotes a diffusion-parameterized conditional distribution
over future visual states. The latent action serves as a shared interface for
visual transitions, enabling OpenNWM to model multimodal future dynamics
without explicit action labels.

\textbf{Discussion.}~
Conventional Navigation World Models rely on action-labeled trajectories, while OpenNWM discovers and grounds latent actions from heterogeneous visual experiences, enabling unified learning from action-free videos and robot data.

\subsection{Universal Navigation Latent Actions}
\label{sec:latent_action}
Visual transitions may contain a mixture of agent motion, environmental dynamics, and observation variations. Latent actions learned solely from visual prediction may capture arbitrary changes rather than meaningful agent behaviors. We address this ambiguity by learning a continuous latent action representation that captures visual transitions while preserving controllable motion semantics.

\textbf{Variational Transition Modeling.}~
Given an initial observation $x_{\tau}$ and a future observation $x_m$, we infer a continuous latent action representation $z_{\tau\rightarrow m}$ that captures the corresponding
visual transition. The resulting variational objective is formulated as:
\begin{equation}
\label{eq:visual_latent_objective}
\mathcal{L}_{\mathrm{vis}}
=
-\mathbb{E}_{q_{\phi}}
\left[
\log p_{\psi}
(x_m\mid x_\tau,z_{\tau\rightarrow m})
\right]
+
\beta D_{\mathrm{KL}}
\left(
q_{\phi}(z_{\tau\rightarrow m}\mid x_\tau,x_m)
\|p(z)
\right)
\end{equation}
Here, $q_\phi$ denotes a diagonal Gaussian posterior, $p(z)=\mathcal{N}(0,I)$ is the latent prior, and $p_\psi$ is an auxiliary transition predictor. The reconstruction term preserves transition information, while KL regularization constrains the latent space. The predictor only regularizes latent action learning, with high-fidelity generation handled by the Navigation World Model.

\textbf{Physical Motion Grounding.}~
Visual transitions may capture environmental and observation changes beyond controllable agent motion. For trajectories with physical annotations, we align latent actions with motion transitions and robot actions through the following objective:
\begin{equation}
\label{eq:physical_grounding}
\mathcal{L}_{\mathrm{phy}}
=
-\mathbb{E}_{q_{\phi}}
\left[
\log p_{\omega}
\left(
\mathbf{y}_{\tau:m}
\mid
z_{\tau\rightarrow m}
\right)
\right]
\end{equation}
Here, $\mathbf{y}{\tau:m}$ contains available future motion states. The predictor $p_{\omega}$ maps latent actions to physical transitions. This objective is applied only to trajectories with available physical supervision.

\textbf{Heterogeneous Supervision.}~
We learn the latent-action representation from all visual observations, while
applying physical-motion supervision only to trajectories with available
annotations. The resulting joint objective is formulated as:
\begin{equation}
\label{eq:latent_action_objective}
\mathcal{L}_{\mathrm{latent}}
=
\mathbb{E}_{(\mathbf{x},\mathbf{y})\sim\mathcal{D}}
\left[
\mathcal{L}_{\mathrm{vis}}
\right]
+
\lambda_{\mathrm{phy}}
\mathbb{E}_{(\mathbf{x},\mathbf{y})\sim\mathcal{D}_{\mathrm{phy}}}
\left[
\mathcal{L}_{\mathrm{phy}}
\right]
\end{equation}
Here, $\mathcal{D}_{\mathrm{phy}}\subseteq\mathcal{D}$ denotes the subset with
available physical motion annotations, and $\lambda_{\mathrm{phy}}$ controls
the contribution of physical supervision. The objective is applied to all observation sequences, while the physical objective is activated for motion-annotated sequences to ground latent actions in navigation dynamics. The learned latent actions serve as unified conditioning variables for OpenNWM.

\subsection{Latent-Conditioned Navigation World Modeling}
\label{sec:latent_nwm}
Given the learned universal latent-action space, the posterior distribution
$q_{\phi}$ infers latent actions from both action-free videos and motion-annotated trajectories.
Each observation $x_i$ is encoded into a visual latent state $s_i$, and the
historical context at time $\tau$ is denoted by $\mathbf{s}_{0:\tau}$.
We model the conditional distribution of future visual states given historical
context and latent actions (see Figure~\ref{fig:framework}).

\textbf{Latent Conditioning.}~
For a transition from $\tau$ to $m$, we combine the embeddings of the latent action, temporal horizon, and diffusion timestep to construct the conditioning
representation:
\begin{equation}
\label{eq:latent_condition}
\xi_{\tau:m}^{(t)}
=
\psi_z(z_{\tau\rightarrow m})
+
\psi_k(k)
+
\psi_t(t)
\end{equation}
Here, $\psi_z$, $\psi_k$, and $\psi_t$ denote the embedding functions for the latent action, temporal horizon, and diffusion timestep, respectively. The conditioning representation $\xi_{\tau:m}^{(t)}$ is injected through adaptive layer normalization, while historical visual states $\mathbf{s}_{0:\tau}$ are incorporated via cross-attention.

\textbf{Conditional Diffusion Dynamics.}~
At diffusion timestep $t$, the forward process gradually perturbs the target future state $s_m$ with Gaussian noise according to a predefined noise
schedule:
\begin{equation}
\label{eq:forward_diffusion}
s_m^{(t)}
=
\sqrt{\alpha_t}s_m
+
\sqrt{1-\alpha_t}\epsilon,
\qquad
\epsilon\sim\mathcal{N}(0,I)
\end{equation}
Here, $s_m^{(t)}$ denotes the noisy future state at diffusion timestep $t$, and $\alpha_t$ is determined by the noise schedule. Conditioned on the historical context $\mathbf{s}_{0:\tau}$ and the latent-action conditioning
representation $\xi_{\tau:m}^{(t)}$, the denoising model predicts the future state:
\[
\widehat{s}_m =
F_\theta
(s_m^{(t)}
\mid
\mathbf{s}_{\tau-h+1:\tau},
\xi_{\tau:m}^{(t)})
\]
Iterating the reverse diffusion process produces a multimodal distribution of
future visual states conditioned on the inferred latent action.

\textbf{Unified World-Model Learning.}~
OpenNWM is optimized using a hybrid objective combining future-state prediction and a variational diffusion objective:
\begin{equation}
\label{eq:latent_nwm_objective}
\mathcal{L}_{\mathrm{NWM}}
=
\mathbb{E}
\left[
\left\|
s_m-\widehat{s}_m
\right\|_2^2
\right]
+
\lambda_{\mathrm{vlb}}
\mathcal{L}_{\mathrm{vlb}}
\end{equation}
Here, the first term supervises future-state prediction, while
$\mathcal{L}_{\mathrm{vlb}}$ denotes the variational lower bound for learning
the reverse diffusion model, and $\lambda_{\mathrm{vlb}}$ controls its
contribution.
Since latent actions are available for all visual transitions, the same
diffusion objective applies to both action-free and motion-annotated data. Physical annotations only ground the latent-action space, while OpenNWM generates multimodal future states for navigation planning.

\subsection{Navigation Planning}
\label{sec:latent_planning}

Given the visual context $\mathbf{s}_{\tau-h+1:\tau}$ and goal observation
$s^{g}$, the post-trained OpenNWM evaluates candidate real waypoint-action
sequences through action-conditioned future visual rollouts.
We retain energy-based receding-horizon planning while optimizing
directly over real waypoint actions.
Latent actions are used for world-model pretraining rather than as
online planning variables.

\textbf{Action-Conditioned Trajectory Optimization.}~
Let $\mathbf{a}=(a_0,\ldots,a_{H-1})$ denote a candidate real
waypoint-action sequence over a planning horizon $H$.
Conditioned on the visual context and each candidate action sequence,
OpenNWM rolls out the corresponding future visual states
$\widehat{\mathbf{s}}=(\widehat{s}_1,\ldots,\widehat{s}_H)$.
We evaluate each candidate rollout using the following planning energy:
\begin{equation}
\label{eq:planning_energy}
\mathcal{E}(\widehat{\mathbf{s}},\mathbf{a};s^g,c)
=
-\mathcal{S}_{\mathrm{goal}}(\widehat{s}_H,s^g)
+
\mathcal{C}_{\mathrm{nav}}(\widehat{\mathbf{s}},\mathbf{a},c).
\end{equation}
Here, $\mathcal{S}_{\mathrm{goal}}$ measures the similarity between the
predicted terminal state and the goal observation, while
$\mathcal{C}_{\mathrm{nav}}$ aggregates navigation constraints,
including predicted-state safety and physical-action feasibility
under the embodiment context $c$.
The optimal action sequence is selected by minimizing the expected
trajectory energy:
\begin{equation}
\label{eq:latent_planning}
\mathbf{a}^{*}
=
\arg\min_{\mathbf{a}}
\mathbb{E}_{\widehat{\mathbf{s}}\sim
p_{\theta}(\cdot\mid\mathbf{s}_{0:\tau},\mathbf{a})}
\left[
\mathcal{E}(\widehat{\mathbf{s}},\mathbf{a};s^g,c)
\right].
\end{equation}
Here, $p_{\theta}(\cdot\mid\mathbf{s}_{0:\tau},\mathbf{a})$
denotes the rollout distribution induced by the post-trained,
real-action-conditioned world model, with the fixed temporal offset
per planning step omitted for brevity.
We solve this optimization using the cross-entropy method (CEM)~\citep{cem},
iteratively updating a proposal distribution toward elite
action sequences evaluated through world-model rollouts.

\textbf{Action Execution.}~
The first optimized action $a_0^{*}$ is executed by the robot.
The auxiliary physical predictor $p_{\omega}$ is used only for
motion grounding during latent-action learning and is not required
for online planning.
Upon receiving a new observation, OpenNWM updates its visual context
and repeats action-space optimization.
This receding-horizon procedure enables predictive planning directly
in the real waypoint-action space.

\input{tables/algorithm}

\textbf{Action-Conditioned Planning.}~
OpenNWM solves goal-conditioned navigation via receding-horizon
optimization over real waypoint-action sequences.
Candidates are sampled from a Gaussian proposal distribution and
evaluated through real-action-conditioned world-model rollouts.
Their energies jointly account for goal similarity, predicted-state
safety, and physical-action feasibility.
The lowest-energy candidates are used to update the proposal
distribution over $J$ CEM iterations.
Only the first action of the selected sequence is passed to the robot
controller before replanning from updated observations. Pseudocode is provided in Algorithm~\ref{alg:latent_planning}.
We use $H=8$ planning steps, corresponding to 2 seconds with
0.25-second intervals, and evaluate $N=80$ candidate sequences
with $M=3$ stochastic rollouts per candidate in each CEM iteration.

\subsection{NavAnyWhere}
We construct NavAnywhere, a large-scale heterogeneous visual navigation corpus, by integrating 15 open-source datasets. It contains 17.5M frames of observations, spanning action-free videos across diverse environments. NavAnywhere enables OpenNWM to
jointly learn scalable visual dynamics and physically grounded latent actions.
NavAnywhere details are in
Appendix~\ref{app:navanywhere}; preprocessing and dataset usage are described in Appendix~\ref{app:dataset_details_preprocessing}.

%% file: tables/algorithm.tex

\begin{algorithm}[t]
\caption{Action-Conditioned Receding-Horizon Planning}
\label{alg:latent_planning}

\begin{algorithmic}[1]
\Require Post-trained world model $p_\theta$, visual encoder $E$,
goal observation $x^g$, context $c$
\Require Horizon $H$, context length $h$, candidates $N$, elites $K$,
rollouts $M$, CEM iterations $J$

\State $s^g \gets E(x^g)$

\While{the goal is not reached}
    \State Observe $x_\tau$ and update the encoded visual context
    $\mathbf{s}_{\tau-h+1:\tau}$ using $E$
    \State Initialize
    $\mathcal{Q}_0 =
    \mathcal{N}(\boldsymbol{\mu}_0,\boldsymbol{\Sigma}_0)$

    \For{$j=0,\ldots,J-1$}
        \For{$n=1,\ldots,N$}
            \State Sample
            $\mathbf{a}^{(n)}\sim\mathcal{Q}_j$,
            \quad $\overline{\mathcal{E}}_n\gets0$

            \For{$r=1,\ldots,M$}
                \State $\widehat{\mathbf{s}}^{(n,r)}
                \sim p_\theta\!\left(
                \cdot\mid
                \mathbf{s}_{\tau-h+1:\tau},\mathbf{a}^{(n)}
                \right)$
                \State $\overline{\mathcal{E}}_n\gets
                \overline{\mathcal{E}}_n+
                \frac{1}{M}\mathcal{E}\!\left(
                \widehat{\mathbf{s}}^{(n,r)},
                \mathbf{a}^{(n)};s^g,c
                \right)$
            \EndFor
        \EndFor

        \State $\mathcal{I}_{\mathrm{elite}}\gets
        \operatorname{arg\,sort}_{K}
        \{\overline{\mathcal{E}}_n\}_{n=1}^{N}$

        \State Refit $\mathcal{Q}_{j+1}$ using
        $\{\mathbf{a}^{(n)}:
        n\in\mathcal{I}_{\mathrm{elite}}\}$
    \EndFor

    \State $n^\star\gets
    \arg\min_{n\in\{1,\ldots,N\}}
    \overline{\mathcal{E}}_n$
    \State $\mathbf{a}^\star\gets\mathbf{a}^{(n^\star)}$
    \State $a_\tau^\star\gets a_0^\star$
    \State Execute $a_\tau^\star$ through the robot controller
    and advance to the next observation
\EndWhile

\end{algorithmic}
\end{algorithm}

%% file: main/Experiments.tex
\section{Experiments}
\begin{figure*}[t]
    \centering
     \includegraphics[width=0.88\textwidth]{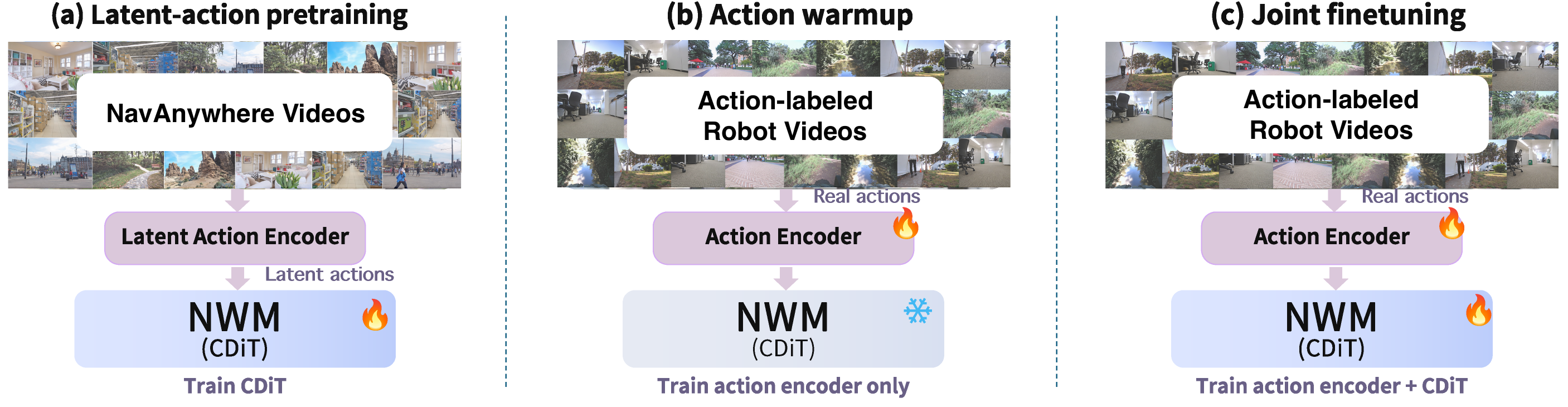}
    \caption{{Three-stage training pipeline of OpenNWM.}
(a) Latent-action pretraining on action-free NavAnywhere videos.
(b) Action-encoder warmup with the CDiT backbone frozen.
(c) Joint post-training of the entire model.}
    \vspace{-6pt}
    \label{fig:three-stage}
\end{figure*}

\subsection{Experiment Setups}
\textbf{Datasets.}~
We pretrain OpenNWM on \textsc{NavAnywhere}, a large-scale action-free video
corpus covering diverse indoor, outdoor, urban, and natural environments.
For downstream training, we use SCAND~\citep{scand}, TartanDrive~\citep{tartandrive}, RECON~\cite{recon}, and HuRoN (SACSoN)~\citep{huron}, whose
pose annotations are converted into relative translations and rotations.
Translational displacements are normalized by the average step length of each
dataset, and backward-motion transitions are removed. We evaluate in-domain performance on these four datasets and assess generalization to unseen environments on Go Stanford, the TUM RGB-D SLAM Dataset, Office-Go2, and Planetary Rover. Office-Go2 contains real-world office navigation sequences that we collected using a quadruped robot, while Planetary Rover comprises lunar and Martian rover navigation data sourced from the Internet. Details are provided in the Appendix~\ref{app:ood_datasets}.

\textbf{Evaluation Metrics.}~
ATE and RPE assess trajectory accuracy and local pose consistency~\citep{sturm2012benchmark}.
PSNR, LPIPS~\citep{zhang2018unreasonable}, and
DreamSim~\citep{fu2023dreamsim} measure visual fidelity,
while FID~\citep{heusel2017gans} assesses image distribution quality.
We also evaluate LAM reconstruction under ID and OOD settings.
Details are provided in the Appendix~\ref{app:evaluation_metrics}

\textbf{Baselines.}~
We compare OpenNWM with NWM~\citep{bar2025navigation} and RAE-NWM~\citep{zhang2026rae} for visual prediction and planning, and include NWM CDiT-XL + Ego4D~\citep{grauman2022ego4d} for visual prediction. Navigation baselines additionally include GNM~\citep{shah2023gnm}, NoMaD~\citep{sridhar2024nomad}, CompACT~\citep{kim2026planning}, and the latent-action controller of \citet{garrido2026learning}.

\textbf{Implementation Details.}~
As shown in Figure~\ref{fig:three-stage}. we train the LAM on NavAnywhere together
with motion-annotated trajectories from RECON, HuRoN, and SCAND,
and use its inferred latent actions to pretrain the world model.
We then post-train on four navigation datasets with camera poses
for physical grounding.
OpenNWM uses a CDiT-B/2 backbone and a Stable Diffusion
VAE~\citep{blattmann2023stable}, with four context frames
and four sampled prediction horizons per trajectory.
Both LAM and NWM are trained on 8 NVIDIA L20 GPUs.

\subsection{Comparison Results}

\input{tables/main_results}

\textbf{Visual Prediction.}
Table~\ref{tab:reconstruction_results} reports ID averages over RECON,
SCAND, HuRoN, and TartanDrive, and OOD PSNR on Go Stanford (Stanford),
TUM RGB-D (TUM), Planetary Rover (Rover), and Office Go2 (Go2). OpenNWM achieves the best results among the compared methods on all three metrics.
Relative to the strongest baseline for each metric, it reduces LPIPS and DreamSim by 9.1\% and 13.7\%, respectively, and improves PSNR by 0.342\,dB.
These gains are obtained with a 280M-parameter world model, comparable in size to NWM (279M) and substantially smaller than RAE-NWM (867M) and NWM CDiT-XL + Ego4D (1,099M).
These results suggest that OpenNWM learns a more accurate model of navigation dynamics, enabling more faithful action-conditioned predictions of future observations.

\noindent\textbf{Navigation Performance.}
Table~\ref{tab:navigation_results} evaluates navigation performance on
RECON, SCAND, and Office Go2 using ATE and RPE.
OpenNWM achieves the lowest ATE and the best or tied-best RPE
across all three datasets.
On the in-domain benchmarks, it reduces ATE from 1.13 to 1.12
on RECON and from 1.01 to 0.98 on SCAND relative to NWM.
It also ties with NWM for the best RPE of 0.35 on RECON and
achieves the lowest RPE of 0.27 on SCAND, compared with 0.28
for both NWM and RAE-NWM.
Together with the visual prediction results, these findings
support the effectiveness of OpenNWM as a predictive model
for navigation planning.

\noindent\textbf{Out-of-Distribution Generalization.}
We evaluate generalization to unseen domains through both
visual prediction and navigation planning.
For visual prediction, OpenNWM achieves the highest PSNR on
three of the four OOD datasets in Table~\ref{tab:reconstruction_results},
outperforming the strongest baseline on Go Stanford, TUM RGB-D,
and Planetary Rover.
On Office Go2, it ranks second with a PSNR of 11.047\,dB,
only 0.036\,dB below NWM CDiT-XL + Ego4D.
Beyond visual prediction, Office Go2 also serves as an OOD
navigation benchmark in Table~\ref{tab:navigation_results}.
OpenNWM achieves the lowest ATE of 2.89 and RPE of 0.65,
improving upon the strongest navigation baseline RAE-NWM on this dataset.
Notably, OpenNWM achieves the best reported navigation accuracy
on Office Go2 despite ranking second in PSNR, highlighting the
importance of assessing generalization through downstream
planning as well as visual prediction.
Overall, these results support the transferability of OpenNWM
to the evaluated unseen navigation domains.

\textbf{LAM Reconstruction and Generalization.}
As shown in Table~\ref{tab:lam_results}, our LAM achieves the best RGB reconstruction results in both in-domain (ID) and out-of-domain (OOD) settings, demonstrating consistent advantages in pixel fidelity and perceptual similarity. For ID evaluation, our method achieves a PSNR of 21.071 dB and an LPIPS of 0.505, improving upon the strongest baseline for each metric by 2.121 dB and 2.1\%, respectively. For OOD evaluation, it achieves a PSNR of 21.912 dB and an LPIPS of 0.472, outperforming DreamDojo by 1.199 dB and 2.7\%, respectively. These results suggest that the learned latents retain key visual information and generalize to unseen domains, supporting downstream world-model prediction.
\input{tables/lam_results}

\subsection{Ablation Studies}
To assess the effectiveness of our components, we focus on two questions:
(1) Do latent actions benefit navigation world models?
(2) Which LAM supervision best supports navigation world models? We evaluate visual prediction using LPIPS for direct prediction on
Office Go2 at 4\,s and autoregressive prediction on Go Stanford at
4\,FPS up to 4\,s. Due to computational constraints, we use a smaller subset of NavAnywhere
for pretraining in these ablations. Details are provided in the Appendix~\ref{app:training_details}.

\noindent\textbf{Pretraining and Conditioning Ablation.}
Table~\ref{tab:pretraining_ablation} compares different conditioning
signals during pretraining.
Joint time and latent-action conditioning performs best across all six
metrics, reducing SCAND ATE from 1.045 with time conditioning alone to
0.950.
This suggests
that motion information complements the temporal prediction horizon.

\begin{figure*}[t]
    \centering
     \includegraphics[width=0.98\textwidth]{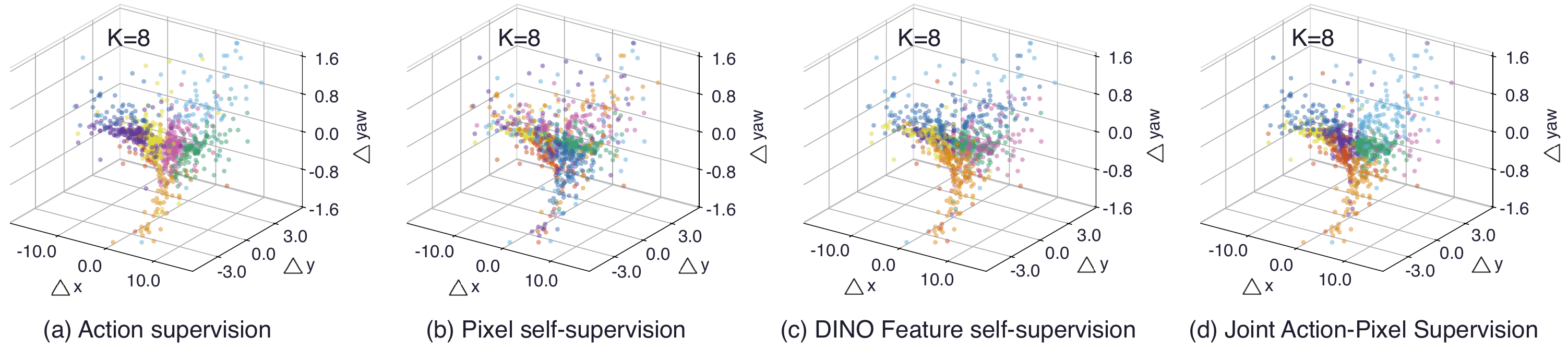}
    \caption{Latent-action clustering under four LAM supervision paradigms.}
    \label{fig:cluster-results}
\end{figure*}

\noindent\textbf{LAM Supervision Ablation.}
Table~\ref{tab:lam_objective_ablation} compares world model performance
using different LAMs learned with pixel-only, action-only, and joint
supervision. As shown in Figure~\ref{fig:matrix}, Pixel+Action supervision achieves the highest Macro-6 Recall on four of five datasets. It consistently outperforms action-only supervision, suggesting that visual reconstruction complements physical motion grounding in aligning latent actions with navigation semantics. Joint supervision leads on five of six metrics.
Figure~\ref{fig:cluster-results} visualizes latent-cluster assignments
in the physical motion space
$(\Delta x,\Delta y,\Delta\mathrm{yaw})$ under four LAM supervision
paradigms.
Compared with pixel-only supervision, joint supervision exhibits more
coherent cluster organization with respect to translation and rotation,
suggesting stronger alignment between latent actions and physical
motion semantics.
Together, these quantitative results and qualitative patterns suggest
that visual prediction and physical-motion grounding provide
complementary supervision for learning effective latent actions.
The Appendix~\ref{app:latentactionanalysis} provides LAM paradigm definitions, qualitative analysis, and semantic-consistency evaluation.

\input{tables/ablation}

\section{Conclusion}
\label{sec:conclusion}
We introduced OpenNWM, a generalist world model for open-world visual navigation. At its core is a shared latent-action interface that unifies learning from heterogeneous visual trajectories with different action spaces and levels of physical supervision. By coupling large-scale action-free video pretraining with physical-motion grounding, OpenNWM learns transition representations that are both predictive of future observations and aligned with executable navigation motion. These representations condition a diffusion world model and support receding-horizon planning after post-training. Experiments across diverse environments demonstrate strong performance in visual future prediction, navigation planning, and generalization to unseen scenes. Taken together, our results establish latent actions as a scalable means of converting open-domain videos into control-relevant navigation supervision, providing a promising foundation for generalist Navigation World Models. Future work will extend this framework to longer-horizon prediction, higher-dimensional action spaces, and online adaptation in dynamic environments.

%% file: tables/main_results.tex
\begin{table}[t]
\centering
\normalsize
\setlength{\tabcolsep}{1pt}

\begin{minipage}[t]{0.54\linewidth}
\vspace{0pt}
\caption{Direct visual prediction at 4\,s.
DS: DreamSim; NWM$^{\dagger}$: CDiT-XL + Ego4D.
ID/OOD denote in-domain/out-of-domain evaluation.
Best and second-best results are \textbf{bold} and
\underline{underlined}, respectively.}
\label{tab:reconstruction_results}
\label{tab:main_results}
\end{minipage}\hfill
\begin{minipage}[t]{0.44\linewidth}
\vspace{0pt}
\caption{Navigation performance on RECON, SCAND,
and Office Go2 (OOD).
Lower ATE and RPE indicate better performance.
Best and second-best results are \textbf{bold} and
\underline{underlined}, respectively.}
\vspace{0pt}
\label{tab:navigation_results}
\end{minipage}


\noindent
\begin{minipage}[t]{0.54\linewidth}
\vspace{0pt}
\centering
\renewcommand{\arraystretch}{1.32}

\resizebox{\linewidth}{!}{%
\begin{tabular}{@{}lr*{7}{c}@{}}
\toprule
\multirow{2}{*}{\textbf{Method}}
& \multirow{2}{*}{\shortstack{\textbf{Params.}\\\textbf{(M)}}}
& \multicolumn{3}{c}{\textbf{ID}}
& \multicolumn{4}{c}{\textbf{OOD} (PSNR$\uparrow$)} \\
\cmidrule(lr){3-5}
\cmidrule(lr){6-9}
&
& LPIPS$\downarrow$
& DS$\downarrow$
& PSNR$\uparrow$
& Stanford
& TUM
& Rover
& Go2 \\
\midrule

\rowcolor[gray]{0.95}
NWM & 279
& \underline{0.405} & 0.230 & 13.811
& 11.323 & 8.957 & 11.739 & 10.750 \\

NWM$^{\dagger}$ & 1,099
& 0.419 & \underline{0.227} & 13.523
& 11.444 & 8.884 & \underline{11.906}
& \textbf{11.083} \\

\rowcolor[gray]{0.95}
RAE-NWM & 867
& 0.442 & 0.254 & \underline{13.957}
& \underline{11.528} & \underline{9.007}
& 9.449 & 10.500 \\

\textbf{OpenNWM} & 280
& \textbf{0.368} & \textbf{0.196} & \textbf{14.299}
& \textbf{11.591} & \textbf{9.135}
& \textbf{12.574} & \underline{11.047} \\

\bottomrule
\end{tabular}%
}
\end{minipage}\hfill
\begin{minipage}[t]{0.44\linewidth}
\vspace{0pt}
\centering
\renewcommand{\arraystretch}{1.0}

\resizebox{\linewidth}{!}{%
\begin{tabular}{@{}lr*{6}{c}@{}}
\toprule
\multirow{2}{*}{\textbf{Method}}
& \multirow{2}{*}{\shortstack{\textbf{Params.}\\\textbf{(M)}}}
& \multicolumn{2}{c}{\textbf{RECON}}
& \multicolumn{2}{c}{\textbf{SCAND}}
& \multicolumn{2}{c}{\textbf{Office Go2}} \\
\cmidrule(lr){3-4}
\cmidrule(lr){5-6}
\cmidrule(lr){7-8}
&
& ATE$\downarrow$
& RPE$\downarrow$
& ATE$\downarrow$
& RPE$\downarrow$
& ATE$\downarrow$
& RPE$\downarrow$ \\
\midrule

\rowcolor[gray]{0.95}
GNM & 9
& 1.87 & 0.73
& 2.12 & 0.61
& -- & -- \\

NoMaD & 19
& 1.95 & 0.53
& 2.24 & 0.49
& -- & -- \\

\rowcolor[gray]{0.95}
NWM & 279
& \underline{1.13} & \textbf{0.35}
& \underline{1.01} & \underline{0.28}
& 3.03 & 0.68 \\

RAE-NWM & 867
& 1.36 & \underline{0.37}
& 1.14 & \underline{0.28}
& \underline{3.01} & \underline{0.67} \\

\rowcolor[gray]{0.95}
\citet{garrido2026learning} & --
& 1.40 & 0.40
& -- & --
& -- & -- \\

CompACT & --
& 1.33 & 0.39
& 1.36 & 0.34
& -- & -- \\

\rowcolor[gray]{0.95}
\textbf{OpenNWM} & 280
& \textbf{1.12} & \textbf{0.35}
& \textbf{0.98} & \textbf{0.27}
& \textbf{2.89} & \textbf{0.65} \\

\bottomrule
\end{tabular}%
}
\end{minipage}

\end{table}

%% file: tables/lam_results.tex
\begin{figure*}[t]
\centering

\begin{minipage}[t]{0.50\textwidth}
\vspace{0pt}
\centering

\captionof{table}{{Image reconstruction performance of LAMs.}
ID (in-domain): average over RECON, SCAND, and HuRoN.
OOD (out-of-domain): unseen Go Stanford.
Best results are in \textbf{bold} and second best are \underline{underlined}.}
\label{tab:lam_results}

\vspace{2pt}

\begingroup

\ifdefined\hypersetup
  \hypersetup{hidelinks}
\fi

\setlength{\tabcolsep}{3pt}
\renewcommand{\arraystretch}{1.1}

\resizebox{\linewidth}{!}{%
\begin{tabular}{@{}llcccc@{}}
\toprule
\textbf{Model}
& \textbf{Source}
& \multicolumn{2}{c}{\textbf{ID}}
& \multicolumn{2}{c}{\textbf{OOD}} \\
\cmidrule(lr){3-4}
\cmidrule(lr){5-6}
&
& \textbf{PSNR} $\uparrow$
& \textbf{LPIPS} $\downarrow$
& \textbf{PSNR} $\uparrow$
& \textbf{LPIPS} $\downarrow$ \\
\midrule

villa-X
& \MainSource{villa-x}
& 15.327
& \underline{0.516}
& 17.560
& 0.550 \\

LARA
& \MainSource{lara}
& 15.752
& 0.538
& 17.742
& 0.527 \\

Olaf-World
& \MainSource{olaf}
& 14.764
& 0.635
& 16.599
& 0.614 \\

DiLA
& \MainSource{dila}
& 13.180
& 0.597
& 14.078
& 0.621 \\

DreamDojo
& \MainSource{dreamdojo}
& \underline{18.950}
& 0.529
& \underline{20.713}
& \underline{0.485} \\

\midrule

\textbf{Ours}
&
& \textbf{21.071}
& \textbf{0.505}
& \textbf{21.912}
& \textbf{0.472} \\

\bottomrule
\end{tabular}%
}

\endgroup
\end{minipage}%
\hfill
\begin{minipage}[t]{0.48\textwidth}
\vspace{0pt}
\centering

\includegraphics[width=\linewidth]{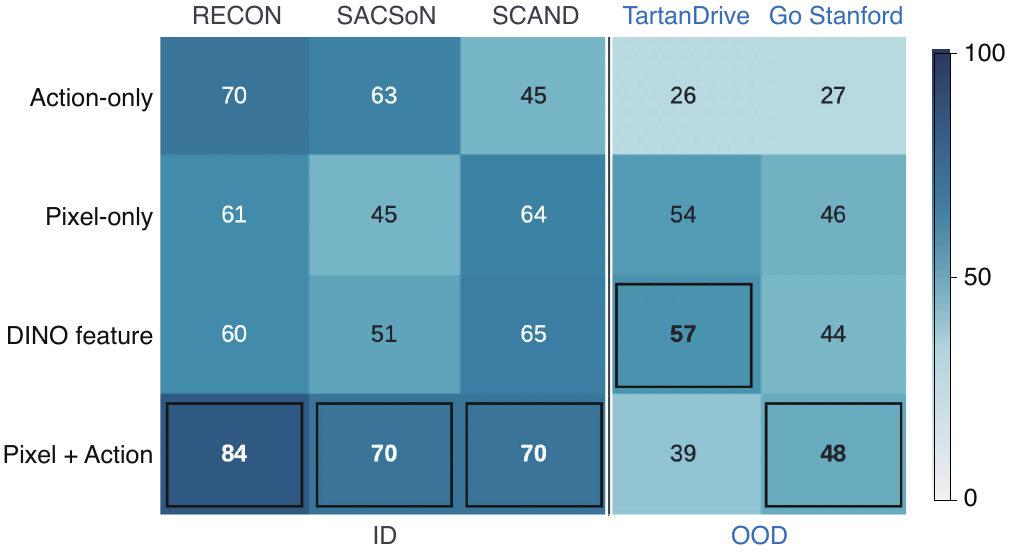}

\captionof{figure}{{Semantic consistency of latent actions.}
Detailed metric definitions are provided in the appendix.}
\label{fig:matrix}

\end{minipage}

\end{figure*}

%% file: tables/ablation.tex
\providecommand{\CompACTCheck}{\ding{51}}
\providecommand{\CompACTCross}{\ding{55}}

\begin{table*}[t]
\centering

\begin{minipage}[t]{0.49\textwidth}
\vspace{0pt}
\centering
\caption{
{Pretraining ablation.}
Effect of different DiT conditioning signals. Best and second-best results are bold and underlined, respectively.
}
\label{tab:pretraining_ablation}

\small
\setlength{\tabcolsep}{1.5pt}
\renewcommand{\arraystretch}{1}

\resizebox{\linewidth}{!}{%
\begin{tabular}{@{}cc|cc|cc|cc@{}}
\toprule
\multicolumn{2}{c|}{\textbf{DiT Conditioning}}
& \multicolumn{2}{c|}{\textbf{Reconstruction}}
& \multicolumn{4}{c}{\textbf{Navigation}} \\

\cmidrule(lr){1-2}
\cmidrule(lr){3-4}
\cmidrule(l){5-8}

Time
& Latent Action
& \textbf{Direct}
& \textbf{Autoregressive}
& \multicolumn{2}{c|}{\textbf{RECON}}
& \multicolumn{2}{c}{\textbf{SCAND}} \\

\cmidrule(lr){3-3}
\cmidrule(lr){4-4}
\cmidrule(lr){5-6}
\cmidrule(l){7-8}

&
& LPIPS $\downarrow$
& LPIPS $\downarrow$
& ATE $\downarrow$
& RPE $\downarrow$
& ATE $\downarrow$
& RPE $\downarrow$ \\

\midrule

\CompACTCross & \CompACTCross
& 0.628 & \underline{0.667}
& 1.198 & 0.367
& 1.252 & 0.316 \\

\CompACTCheck & \CompACTCross
& \underline{0.595} & 0.678
& \underline{1.174} & 0.365
& \underline{1.045} & \underline{0.286} \\

\CompACTCross & \CompACTCheck
& 0.609 & \underline{0.667}
& 1.198 & \underline{0.359}
& 1.067 & 0.289 \\

\CompACTCheck & \CompACTCheck
& \textbf{0.587} & \textbf{0.658}
& \textbf{1.117} & \textbf{0.356}
& \textbf{0.950} & \textbf{0.267} \\

\bottomrule
\end{tabular}%
}
\end{minipage}\hfill%
\begin{minipage}[t]{0.5\textwidth}
\vspace{0pt}
\centering
\caption{
{LAM supervision ablation.}
Effect of different supervision on latent-action learning. Best/second-best results are \textbf{bold}/\underline{underlined}.}
\label{tab:lam_objective_ablation}

\small
\setlength{\tabcolsep}{1pt}
\renewcommand{\arraystretch}{1.1}

\resizebox{\linewidth}{!}{%
\begin{tabular}{@{}cc|cc|cc|cc@{}}
\toprule
\multicolumn{2}{c|}{\textbf{LAM Supervision}}
& \multicolumn{2}{c|}{\textbf{Reconstruction}}
& \multicolumn{4}{c}{\textbf{Navigation}} \\

\cmidrule(lr){1-2}
\cmidrule(lr){3-4}
\cmidrule(l){5-8}

Pixel
& Action
& \textbf{Direct}
& \textbf{Autoregressive}
& \multicolumn{2}{c|}{\textbf{RECON}}
& \multicolumn{2}{c}{\textbf{SCAND}} \\

\cmidrule(lr){3-3}
\cmidrule(lr){4-4}
\cmidrule(lr){5-6}
\cmidrule(l){7-8}

&
& LPIPS $\downarrow$
& LPIPS $\downarrow$
& ATE $\downarrow$
& RPE $\downarrow$
& ATE $\downarrow$
& RPE $\downarrow$ \\

\midrule

\CompACTCross & \CompACTCheck
& \underline{0.598} & \underline{0.670}
& \underline{1.138} & 0.365
& \underline{0.981} & \underline{0.274} \\

\CompACTCheck & \CompACTCross
& 0.605 & 0.674
& 1.176 & \textbf{0.352}
& 0.982 & \underline{0.274} \\

\CompACTCheck & \CompACTCheck
& \textbf{0.587} & \textbf{0.658}
& \textbf{1.117} & \underline{0.356}
& \textbf{0.950} & \textbf{0.267} \\

\bottomrule
\end{tabular}%
}
\end{minipage}

\end{table*}

%% file: main/Appedix.tex
\section{NavAnywhere Dataset Details}
\label{app:navanywhere}

\begin{figure}[H]
    \centering
    \includegraphics[width=0.96\textwidth]{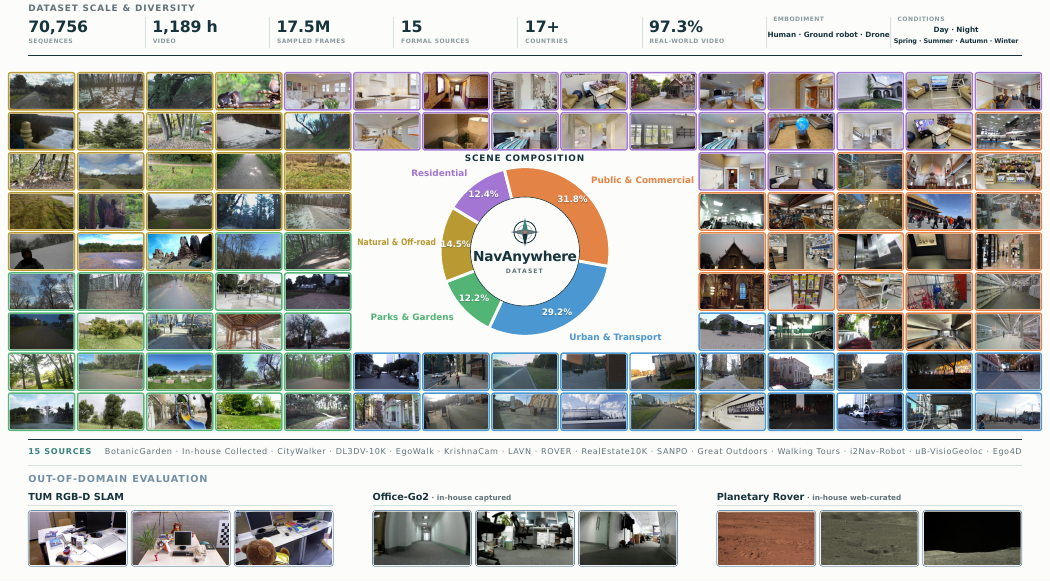}
    \caption{\textbf{Overview of NavAnywhere and its complementary
    out-of-domain evaluation sets.}
    NavAnywhere covers diverse everyday navigation environments.
    The complementary evaluation sets are separate from NavAnywhere
    and are described in Section~\ref{app:ood_datasets}.}
    \vspace{-12pt}
    \label{fig:NavAnywhere}
\end{figure}

We introduce NavAnywhere, a large-scale, action-free visual navigation
dataset assembled from 15 sources, including public datasets and our
in-house recordings. The collection consists of RGB videos and sampled
image sequences without action annotations, comprising 70,756 sequences,
17,496,570 sampled frames, and 1,188.65 hours of source video.
NavAnywhere covers a broad spectrum of environments encountered in
everyday life, ranging from residential and public indoor spaces to
urban streets, parks, and natural outdoor environments. (see Figure~\ref{fig:NavAnywhere})

\subsection{Dataset Composition}
\label{app:navanywhere_composition}

NavAnywhere brings together 15 sources, including public datasets and in-house recordings, covering a broad range of navigation environments, camera viewpoints, and motion patterns. Below, we briefly describe each source and its main characteristics to provide an overview of the dataset’s composition:

\begin{itemize}
    \item \textbf{BotanicGarden~\citep{botanicgarden}:}
    Robot-mounted observations of botanical gardens, featuring
    vegetation, pathways, and other outdoor structures.
    \item \textbf{CityWalker~\cite{liu2025citywalker}:}
    Urban navigation sequences featuring sidewalks, intersections,
    pedestrians, and vehicles.
    \item \textbf{DL3DV-10K~\citep{ling2024dl3dv}:}
    Scene-exploration videos spanning diverse indoor and outdoor
    environments, viewpoints, and camera motions.
    \item \textbf{Ego4D~\cite{grauman2022ego4d}:}
    A navigation-related subset of egocentric videos depicting
    everyday activities and movement through indoor and outdoor
    environments.
    \item \textbf{EgoWalk~\citep{akhtyamov2025egowalk}:}
    First-person walking videos covering urban, residential,
    public, and recreational environments.
    \item \textbf{KrishnaCam~\citep{krishnacam}:}
    Long-term egocentric recordings of daily activities, including
    walking, commuting, and visits to indoor and outdoor spaces.
    \item \textbf{LAVN~\citep{lavn}:}
    A collection combining real-world recordings with simulated
    navigation sequences.
    \item \textbf{ROVER~\citep{rover}:}
    Robot navigation recordings collected in gardens, parks, and
    campus environments across different seasons and illumination
    conditions.
    \item \textbf{RealEstate10K~\citep{realestate10k}:}
Real-estate video clips of residential interiors, exteriors, and surroundings, including drone footage.
    \item \textbf{SANPO~\citep{sanpo}:}
    Real and synthetic egocentric navigation sequences covering
    sidewalks, roads, parks, and other outdoor environments.
    \item \textbf{The Great Outdoors~\citep{go}:}
    Robot-mounted observations of natural and off-road environments,
    including forests, unpaved trails, and uneven terrain.
    \item \textbf{Walking Tours~\citep{walkingtour}:}
    Long-form internet videos depicting walking tours and outdoor
    exploration.
    \item \textbf{i2Nav-Robot~\citep{i2nav}:}
    Robot navigation sequences covering campus buildings, streets,
    parking areas, and transitions between indoor and outdoor spaces.
    \item \textbf{uB-VisioGeoloc~\citep{ub}:}
    Pedestrian-view urban sequences covering streets, public spaces,
    and parks, together with synthetic sequences.
    \item \textbf{In-house collected data:}
    Our in-house collection of nine RGB videos,
    covering offices, residential neighborhoods, and public landmarks
    across indoor and outdoor environments under daytime and
    nighttime conditions.
\end{itemize}

\begin{figure}[t]
    \centering
    \includegraphics[width=0.8\textwidth]{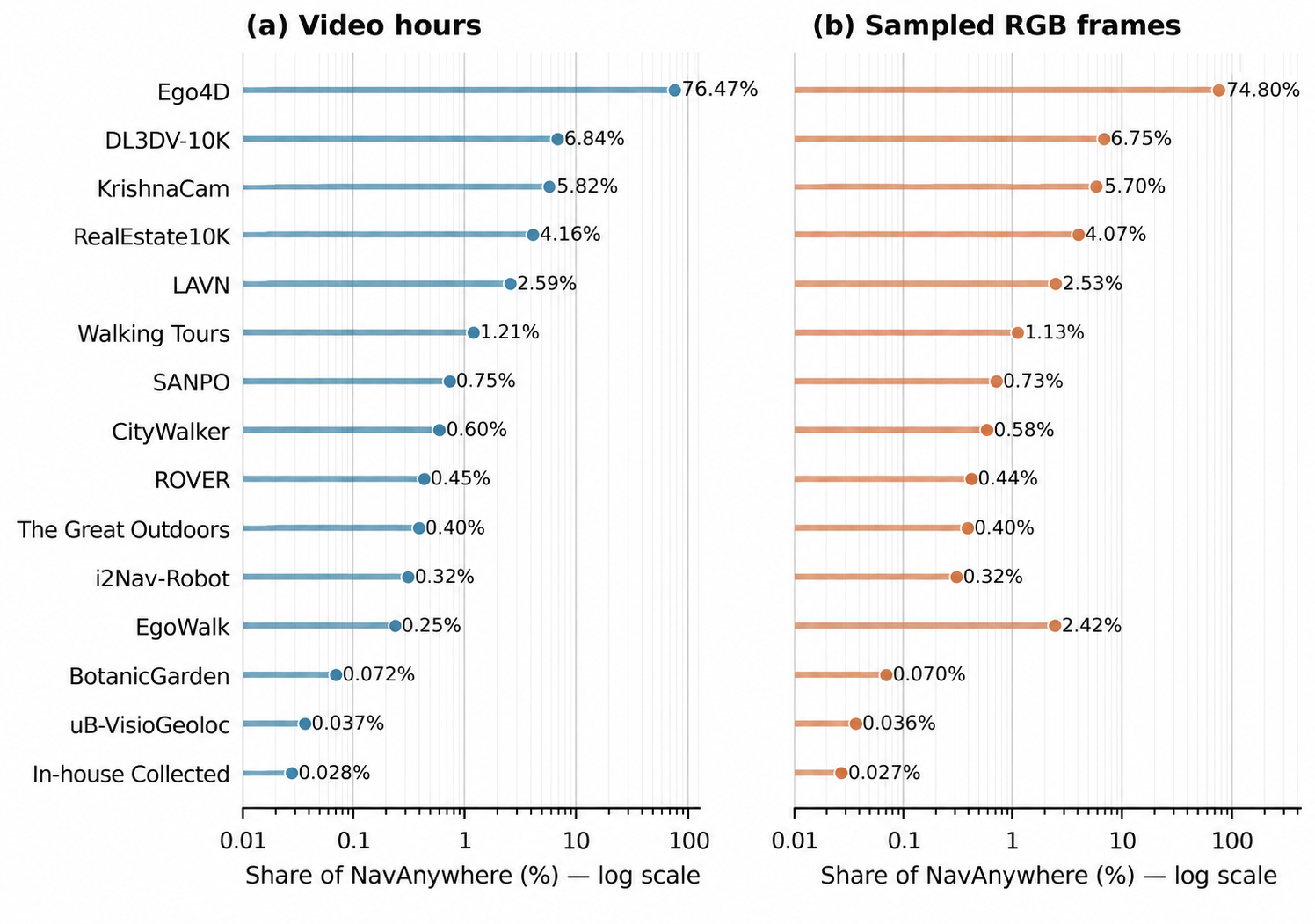}
    \caption{\textbf{Source composition of NavAnywhere.}
    Statistics cover the 15 action-free sources.}
    \vspace{-12pt}
    \label{fig:navanywhere_composition}
\end{figure}

Table~\ref{tab:navanywhere_statistics} and Figure~\ref{fig:navanywhere_composition} reports the complete NavAnywhere dataset statistics. 

\subsection{Scene and Environment Diversity}
\label{app:navanywhere_scene}

NavAnywhere covers a wide range of indoor and outdoor environments, reflecting the diversity of scenes encountered in everyday navigation. We organize these environments into five broad scene categories, each characterized by distinct spatial layouts and environmental features, as described below:

\begin{itemize}
    \item \textbf{Residential environments:}
    Homes, residential buildings, and surrounding neighborhoods.

    \item \textbf{Public and commercial environments:}
    Offices, schools, shops, restaurants, and cultural facilities.

    \item \textbf{Urban and transportation environments:}
    Streets, sidewalks, intersections, parking areas, and transit
    spaces.

    \item \textbf{Parks and managed outdoor environments:}
    Gardens, campuses, recreational areas, and maintained green
    spaces.

    \item \textbf{Natural and off-road environments:}
    Forests, trails, uneven terrain, and unstructured outdoor spaces.
\end{itemize}

Across these categories, the constituent sources exhibit substantial
variation in viewpoints, illumination conditions, scene layouts, and
motion patterns.

\begin{table}[t]
\centering
\caption{\textbf{NavAnywhere dataset statistics.}
Sequences denote processed video clips or trajectories, and frames
denote sampled RGB images. Hours measure source-video duration. All 15 sources are
used as RGB-only, action-free data.}
\label{tab:navanywhere_statistics}

\small
\setlength{\tabcolsep}{24pt}
\resizebox{0.9\linewidth}{!}{
\begin{tabular}{lrrrl}
\toprule
\textbf{Dataset}
& \textbf{Sequences}
& \textbf{Frames}
& \textbf{Hours} \\
\midrule

Ego4D
& 1,619  & 13,088,098 & 908.95 \\
DL3DV-10K
& 20,352 & 1,180,382  & 81.36  \\
KrishnaCam
& 436    & 996,666    & 69.21  \\
RealEstate10K
& 45,010 & 712,145    & 49.44  \\
LAVN
& 312    & 443,244    & 30.78  \\
Walking Tours
& 10     & 198,097    & 14.40  \\
SANPO
& 2,662  & 128,233    & 8.91   \\
CityWalker
& 18     & 102,083    & 7.09   \\
ROVER
& 39     & 76,442     & 5.31   \\
The Great Outdoors
& 17     & 69,162     & 4.80   \\
i2Nav-Robot
& 10     & 55,202     & 3.83   \\
EgoWalk
& 239    & 423,389    & 2.94   \\
BotanicGarden
& 7      & 12,272     & 0.85   \\
uB-VisioGeoloc
& 16     & 6,382      & 0.44   \\
In-house Collected
& 9      & 4,773      & 0.33   \\

\midrule
\textbf{NavAnywhere total}
& \textbf{70,756}
& \textbf{17,496,570}
& \textbf{1,188.65} \\

\bottomrule
\end{tabular}
}
\end{table}

\section{Dataset Details and Preprocessing}
\label{app:dataset_details_preprocessing}

In addition to NavAnywhere, we use five action-annotated
navigation datasets and three complementary out-of-domain evaluation
sets. These datasets are not part of NavAnywhere and are excluded
from its corpus statistics. This section describes these
datasets, the preprocessing procedures applied to all data sources,
and their respective roles in OpenNWM training and evaluation.

\subsection{Action-Annotated Datasets}
\label{app:navanywhere_supervision}
\begin{table}[H]
\centering
\caption{\textbf{Statistics of the action-annotated datasets.}
Sequences denote processed trajectories, and frames denote sampled
RGB images. Go Stanford is used only for evaluation.}
\label{tab:external_dataset_statistics}

\small
\setlength{\tabcolsep}{24pt}
\resizebox{0.9\linewidth}{!}{
\begin{tabular}{lrrl}
\toprule
\textbf{Dataset}
& \textbf{Sequences}
& \textbf{Frames}
& \textbf{Data used} \\
\midrule

RECON
& 11,835 & 610,911 & RGB + position/yaw \\
HuRoN 
& 2,839  & 230,911 & RGB + position/yaw \\
Go Stanford
& 3,696  & 198,126 & RGB + position/yaw \\
SCAND
& 604    & 103,457  & RGB + position/yaw \\
TartanDrive
& 1,251  & 62,884   & RGB + position/yaw \\

\bottomrule
\end{tabular}
}
\end{table}
Following the dataset usage in NWM~\citep{bar2025navigation}, we
separately use RECON, HuRoN, SCAND, TartanDrive, and Go Stanford.
Their processed trajectories provide RGB observations, planar
positions, and yaw, from which relative waypoint actions
$(\Delta x,\Delta y,\Delta\psi)$ are derived in the reference frame
of the current observation.

Go Stanford is reserved exclusively for unseen-environment evaluation.
The other four datasets provide motion supervision according
to the LAM and OpenNWM training protocols described in
Section~\ref{app:navanywhere_usage}.
Table~\ref{tab:external_dataset_statistics} summarizes the processed collections.

\subsection{Complementary Out-of-Domain Evaluation Sets}
\label{app:ood_datasets}

We evaluate out-of-domain transfer on three challenging real-world datasets excluded from training: TUM RGB-D, Planetary Rover, and Office-Go2.

\noindent\textbf{TUM RGB-D.}
We select five sequences from the TUM RGB-D SLAM
benchmark~\citep{sturm2012benchmark}, using RGB observations and the
provided ground-truth camera poses. The processed subset contains
1,189 sampled frames.

\noindent\textbf{Planetary Rover.}
We curate extraterrestrial navigation observations from publicly
available online sources, using released imagery and official pose
metadata from the Chang'e-4/Yutu-2 lunar mission and the
Tianwen-1/Zhurong Martian mission. The collection contains
66 lunar and 173 Martian RGB images, organized into
28 sequences with 211 adjacent transitions whose relative motions
are derived from pose metadata.

\noindent\textbf{Office-Go2.}
We collect navigation recordings in-house using a Unitree Go2
quadruped robot in a real office environment. The collection
includes RGB videos and synchronized camera-pose measurements.
Its seven recordings are processed into 15 sequences containing
2,292 sampled frames.

\subsection{Data Processing}
\label{app:navanywhere_processing}

\noindent\textbf{Temporal sampling.}
We organize sampled
frames in temporal order within each sequence. 

Most NavAnywhere sources are sampled at 4\,Hz, corresponding to a
frame interval of 0.25\,s. EgoWalk, in contrast, is originally recorded
at 100,Hz to capture relatively fast ego-motion. So EgoWalk sequences have an
effective sampling rate of approximately 40\,Hz, corresponding to
a 0.025\,s interval. 

\noindent\textbf{Image-pair sampling for LAM training.}
For LAM training, image pairs are sampled within the same sequence
using signed frame offsets
$\delta \in \{-8,\ldots,8\}$.
These offsets correspond to temporal displacements of up to 2\,s
for 4\,Hz sources and approximately 0.2\,s for EgoWalk.

\noindent\textbf{Context and target sampling for pretraining.}
For pretraining, each example contains four consecutive context
frames and four target frames sampled at signed offsets within
$[-64,64]$ relative to the latest context frame. At 4\,Hz, these
offsets span $[-16,16]$\,s; for EgoWalk, they span approximately
$[-1.6,1.6]$\,s. Latent-action conditioning is available for local
image pairs satisfying $|\delta| \leq 8$.

\noindent\textbf{Processing of action-annotated and evaluation data.}
The five external action-annotated datasets are converted to 4,Hz trajectories with image-aligned planar positions and yaw. TUM RGB-D and Office-Go2 pair sampled RGB frames with camera poses, while Planetary Rover retains the released observation order and endpoint poses without temporal image interpolation.

\subsection{Dataset Usage in OpenNWM}
\label{app:navanywhere_usage}

\noindent\textbf{Action-free learning from NavAnywhere.}
NavAnywhere provides action-free visual observations for latent-action
learning and world-model pretraining. Its training inputs consist
only of RGB observations, even when the original sources provide
additional sensing modalities or annotations. Latent actions are
inferred from image pairs by the learned latent action model.

\noindent\textbf{Physical-motion supervision and evaluation.}
Physical-motion grounding of the LAM uses the training splits of
RECON, HuRoN, and SCAND. Action-conditioned post-training uses
RECON, HuRoN, SCAND, and TartanDrive. Go Stanford and all three complementary
out-of-domain evaluation sets are excluded from these training stages.

\section{Architecture and Implementation Details}
\input{tables/paras}

\subsection{Training and Inference Details}
\label{app:training_details}

Table~\ref{tab:training_details} summarizes the training and
inference configurations of OpenNWM used in the main experiments.

\noindent\textbf{Latent-action learning.}
Our Pixel+Action navigation LAM follows the spatiotemporal
Transformer architecture of~\citet{dreamdojo}. It contains
approximately 709M parameters, comprising a 24-block spatiotemporal
encoder and a 24-block spatial decoder, both with a hidden width
of 1,024 and 16 attention heads. The encoder maps a pair of RGB
frames to a 32-dimensional variational latent. Frames are
center-cropped to a $4{:}3$ aspect ratio and resized to
$240\times320$ (height $\times$ width), with $16\times16$ patches.
Frame pairs use uniformly sampled integer temporal offsets
in $[-8,8]$.

LAM training combines NavAnywhere with
the training splits of RECON, HuRoN, and SCAND, reserving 25\%
of each batch for action-labeled samples. The objective combines
RGB reconstruction MSE, action MSE on labeled samples, and KL
regularization, with respective weights $1$, $0.171244$, and
$10^{-6}$. We train from scratch using AdamW with a global batch
size of 160, a peak learning rate of $2.5\times10^{-5}$, and
weight decay of $0.01$, using BF16 on eight NVIDIA L20 GPUs.
The learning-rate schedule is configured for 100K steps, with
12.8K steps of linear warm-up followed by cosine decay toward
10\% of the peak rate. We use the 60K-step LAM checkpoint to
extract deterministic posterior means as latent actions for pretraining.

\noindent\textbf{World-model pretraining.}
OpenNWM uses a CDiT-B/2 diffusion transformer with approximately
197M parameters and a frozen VAE tokenizer. Images are
resized to $224\times224$. Each observation provides four context
frames and four sampled prediction targets. Pretraining uses
the NavAnywhere subset, with signed target frame
offsets in $[-64,64]$. All samples receive relative-time
conditioning. Valid pairs within $[-8,8]$ additionally receive
their layer-normalized 32-dimensional latent actions;
pairs outside this range contribute to the diffusion objective
without latent-action conditioning.

We use the 60K-step pretraining checkpoint, trained with AdamW
at a learning rate of $2\times10^{-4}$, weight decay of $0.01$,
and gradient clipping at 2.0. Training uses BF16 on eight NVIDIA
L20 GPUs, with 48 observations per GPU, giving a global batch
of 384 observations and 1,536 context-target pairs per update.

\noindent\textbf{World model post-training.}
For post-training, we replace the latent action adapter with
a newly initialized adapter for three-dimensional physical motion
$(\Delta x,\Delta y,\Delta\psi)$. We first train this adapter
for 3K steps with the backbone frozen, followed by 100K steps
of joint post-training of the adapter and backbone. The adapter
learning rate is $10^{-4}$ throughout post-training, and the
backbone learning rate is $10^{-4}$ during the joint phase.
Both phases use AdamW with weight decay of $0.01$ and the same
batch size, precision, and hardware as pretraining.

Post-training uses the training splits of RECON, HuRoN, SCAND,
and TartanDrive. Go Stanford is excluded from training.
Physical-motion targets are expressed relative to the latest
context frame, with translations divided by dataset-specific
waypoint spacing before fixed min--max normalization.

\noindent\textbf{Inference and planning.}
Visual prediction uses 250-step DDPM~\citep{ddpm} sampling with fixed
evaluation windows and seeds. Planning evaluates sequences
of eight physical-motion actions at $0.25\,\mathrm{s}$ intervals,
corresponding to a $2\,\mathrm{s}$ horizon. The cross-entropy
method (CEM) uses 80 candidates, five elites, and one proposal
update. Candidate costs are computed using LPIPS-Alex and
averaged over three stochastic evaluations. Planning conditions
the post-trained world model directly on physical motion.

\noindent\textbf{Ablation settings.}
The ablations retain the same CDiT-B/2
backbone, image resolution, context and target counts, and training data sources as the main experiments. The main difference are less NavAnywhere data to train the LAM, 200K pretraining steps at a learning rate of
$10^{-4}$, and a global batch of 128 observations. Post-training uses a 10K-step adapter
warm-up followed by 100K joint updates, with adapter and
backbone learning rates of $10^{-4}$ and $10^{-5}$,
respectively. The ablations use the same 250-step DDPM and
CEM80 planning protocol described above.

\subsection{Evaluation Metrics}
\label{app:evaluation_metrics}

We evaluate visual prediction quality using three metrics:
Peak Signal-to-Noise Ratio (PSNR), Learned Perceptual Image Patch Similarity (LPIPS), and
Fr\'echet Inception Distance (FID), covering pixel-level fidelity,
structural consistency, perceptual similarity, and distributional
similarity.

\paragraph{Pixel-Level Fidelity.}
Peak Signal-to-Noise Ratio (PSNR) measures reconstruction fidelity
on a logarithmic scale. Given the predicted image $\hat{\mathbf I}$
and the ground-truth image $\mathbf I$, it is defined as:
\begin{equation}
\mathrm{PSNR}
=
10\log_{10}
\left(
\frac{I_{\max}^{2}}
{\frac{1}{N}\sum_{i=1}^{N}(\hat I_i-I_i)^2}
\right),
\label{eq:psnr}
\end{equation}
where $N$ denotes the total number of image elements, including all
spatial locations and color channels, and $I_{\max}$ denotes the
maximum possible intensity value under the adopted image
representation. PSNR is expressed in decibels (dB), with higher
values indicating higher pixel-level fidelity.

\paragraph{Perceptual Similarity.}
Learned Perceptual Image Patch Similarity (LPIPS) measures
perceptual discrepancy in the feature space of a pretrained
vision network. Given channel-normalized feature representations
$\phi_l(\cdot)$ at layer $l$, it is defined as:
\begin{equation}
\mathrm{LPIPS}(\hat{\mathbf I},\mathbf I)
=
\sum_l
\frac{1}{H_lW_l}
\sum_{h,w}
\left\|
\mathbf w_l \odot
\left(
\phi_l(\hat{\mathbf I})_{hw}
-
\phi_l(\mathbf I)_{hw}
\right)
\right\|_2^2,
\label{eq:lpips}
\end{equation}
where $\mathbf w_l$ denotes the learned channel-wise weights,
$\odot$ denotes element-wise multiplication, and
$H_l\times W_l$ is the spatial resolution of the corresponding
feature map. Lower LPIPS indicates higher perceptual similarity,
complementing pixel-level and structural measures.

\paragraph{Distributional Similarity.}
Fr\'echet Inception Distance (FID)~\citep{heusel2017gans} measures
the discrepancy between the feature distributions of predicted
and ground-truth image sets. Features are extracted using a
pretrained Inception-v3 network, and each feature distribution
is approximated by a multivariate Gaussian.
Let $(\boldsymbol{\mu}_p,\boldsymbol{\Sigma}_p)$ and
$(\boldsymbol{\mu}_r,\boldsymbol{\Sigma}_r)$ denote the empirical
means and covariance matrices of the predicted and ground-truth
image features, respectively. FID is defined as:
\begin{equation}
\mathrm{FID}
=
\left\|
\boldsymbol{\mu}_p-\boldsymbol{\mu}_r
\right\|_2^2
+
\operatorname{Tr}
\left(
\boldsymbol{\Sigma}_p
+
\boldsymbol{\Sigma}_r
-
2
\left(
\boldsymbol{\Sigma}_p\boldsymbol{\Sigma}_r
\right)^{1/2}
\right),
\label{eq:fid}
\end{equation}
where $\operatorname{Tr}(\cdot)$ denotes the matrix trace and
$(\cdot)^{1/2}$ denotes the matrix square root.
Lower FID indicates closer agreement between the feature
distributions of predicted and ground-truth images.
Unlike PSNR and LPIPS, FID is computed over image sets
rather than individual image pairs and does not directly
measure prediction--target correspondence.

Together, these four metrics provide complementary assessments
of visual prediction quality, from paired reconstruction
fidelity to distribution-level similarity.

\section{Extended Quantitative Results}
\label{app:extended_quantitative_results}

\noindent\textbf{Consistency across Datasets.}
Table~\ref{tab:id_direct_4s} shows that OpenNWM improves over NWM
on every visual prediction metric across all four in-domain datasets.
In comparison, RAE-NWM achieves stronger results on HuRoN but
underperforms NWM on TartanDrive, revealing
greater variation in its relative performance across datasets.
The advantage of OpenNWM therefore extends beyond an improvement
in the average score: its gains are distributed across the
evaluated sources without degrading any reported dataset--metric
combination relative to NWM.
This consistency also extends to navigation
(Table~\ref{tab:main_navigation}), where OpenNWM reduces ATE
on every dataset while matching or improving RPE.

\begin{figure}[t]
  \centering
  \includegraphics[width=0.95\linewidth]{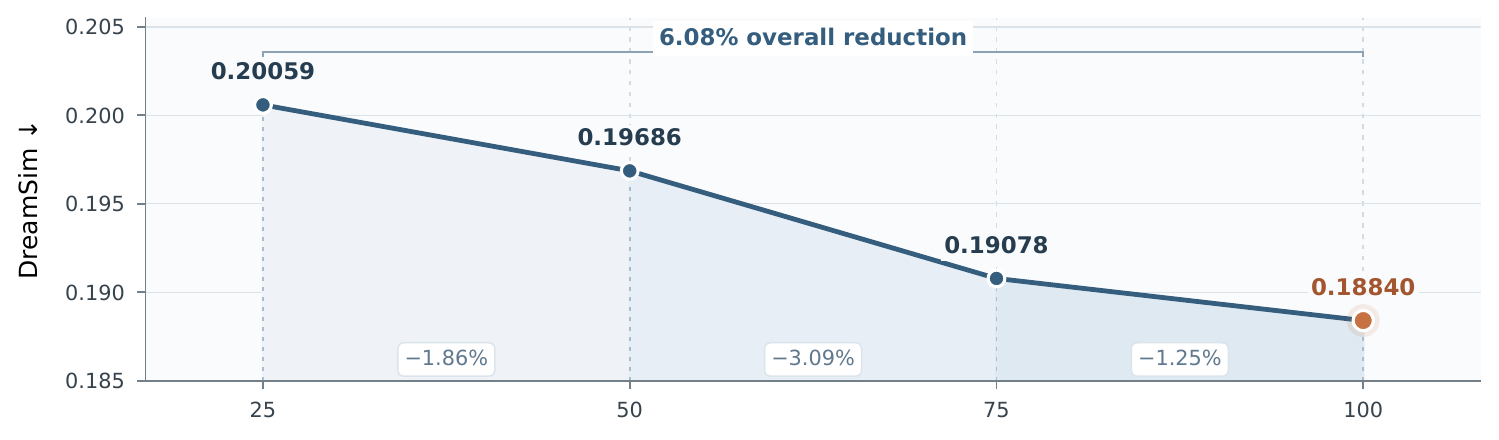}
  \caption{\textbf{Scaling unlabeled data ablation.}
  The horizontal axis gives the percentage of the unlabeled data pool
  used for latent-action pretraining, with action-label availability
  fixed.}
  \label{fig:data_scaling}
\end{figure}
\input{tables/appendix_id}

\noindent\textbf{Visual Fidelity and Navigation Utility.}
SCAND reveals a mismatch between visual prediction and navigation
rankings: RAE-NWM achieves better PSNR and DreamSim than OpenNWM,
yet OpenNWM obtains lower ATE and RPE.
Moreover, the visual metrics themselves disagree, with LPIPS
favoring OpenNWM.
These results indicate that individual image-similarity metrics
are incomplete proxies for downstream navigation performance.
Direct prediction measures agreement with an observed future,
but does not directly assess whether predicted transitions support
effective action selection.
The navigation results therefore provide complementary evidence
of the model's utility for control.

\section{Scaling and Ablation Studies}
\label{sec:scaling_ablation}

We examine how the amount of unlabeled video used for latent-action
learning affects the navigation world model. For this ablation, we
compare four latent-action pretraining settings that retain 25\%, 50\%,
75\%, or 100\% of the unlabeled data pool. We denote this percentage by
$U$ and keep the action-label availability fixed.
Each resulting model is then fine-tuned with real actions using the same
downstream architecture, hyperparameters, and fine-tuning schedule.
We evaluate all four models on the same 500 RECON windows for direct
4\,s prediction. We report the mean DreamSim over these
predictions; lower values indicate greater perceptual similarity to
the ground-truth images.

Figure~\ref{fig:data_scaling} shows a monotonic decrease in RECON
DreamSim at $U=25\%$, $50\%$,
$75\%$, and $100\%$, respectively. Using the full unlabeled data pool
reduces DreamSim by 6.08\% relative to the 25\% setting. The improvement
continues from 75\% to 100\%. This supports the benefit of
scaling unlabeled video for perceptual prediction quality.

\section{Latent-Action Visualization and Analysis}
\label{app:latentactionanalysis}
We further analyze whether learned Universal Navigation Latent Actions
(UNLAs) capture physically grounded navigation semantics and generalize
across heterogeneous environments. Specifically, we investigate two aspects:
(1) physical alignment between latent actions and induced motions, and
(2) cross-dataset consistency of latent action semantics.

\input{tables/appendix_navigation}

\subsection{Latent action learning paradigms.}
\begin{figure*}[t]
    \centering
    \includegraphics[width=0.9\textwidth]{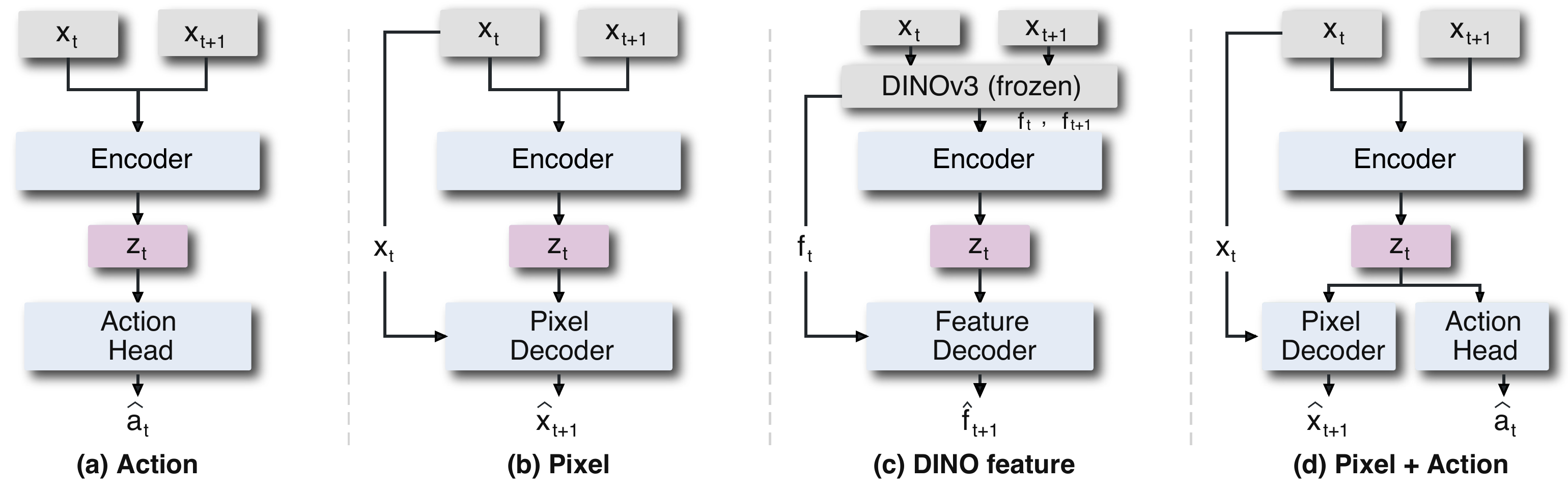}
    \caption{Four latent action learning paradigms.
    An encoder maps a visual transition to a variational latent $z_t$.
    The prediction targets are (a) the observed action $a_t$, (b) the future image $x_{t+1}$,
    (c) frozen DINOv3 features $f_{t+1}$, and (d) both the image and action.
    Reconstruction decoders additionally receive the current image or its features.
    In (d), pixel reconstruction uses all samples, while the action loss uses only labeled samples.
    All variants include the same form of KL regularization.}
    \label{fig:lam-paradigms}
\end{figure*}

As shown in Figure~\ref{fig:lam-paradigms}, we compare four paradigms for learning continuous latent actions from visual transitions $(x_t,x_{t+1})$.
Each infers a variational latent $z_t$ and applies KL regularization toward a standard Gaussian prior.
\textbf{Action} predicts the observed navigation action $a_t$ from $z_t$ using an action head.
\textbf{Pixel} reconstructs $x_{t+1}$ conditioned on $(x_t,z_t)$.
\textbf{DINO feature} operates in a frozen DINOv3~\citep{dinov3} feature space: its encoder infers $z_t$ from $(f_t,f_{t+1})$, where $f_t=\phi(x_t)$, and its decoder predicts $f_{t+1}$ from $(f_t,z_t)$.
\textbf{Pixel + Action} jointly optimizes pixel reconstruction and action prediction from the same latent, applying action supervision only to labeled samples.
All prediction losses use mean squared error; the joint objective weights the action term by $\lambda$.
These paradigms examine how action supervision, visual reconstruction, and their combination shape the learned representation.

\subsection{Physical Alignment of Latent Actions}
\label{sec:physical_alignment}

To investigate whether latent actions encode meaningful navigation dynamics, we analyze the correspondence between latent actions and their associated physical motions. Given an anchor observation and a target observation, each model generates a 32-dimensional latent action representation. For each supervision paradigm, we independently perform K-means clustering in the original 32-dimensional latent space (rather than the projected visualization space) with $K=8$, obtaining eight latent action clusters
$C_1,\ldots,C_8$.

Importantly, clustering is performed solely in the original latent space,
rather than the three-dimensional motion space used for visualization. After clustering, each sample is visualized in the physical motion space
$(\Delta x,\Delta y,\Delta yaw)$, where each point represents an
anchor-target observation pair and the color indicates its latent cluster
assignment.

Compact clusters in the physical space indicate that samples sharing the same
latent action cluster correspond to similar navigation motions. Conversely, highly overlapping clusters suggest that the latent action representation is affected by non-motion factors such as scene appearance,
texture, or observation variations.

Figure~\ref{fig:cluster-results} shows the clustering results under different supervision paradigms. Action supervision produces the most
compact and clearly separated motion clusters, with different clusters aligning well with translational and rotational motion patterns. Joint Action-Pixel supervision achieves similar motion-oriented structures, suggesting that visual supervision improves representation richness while
physical supervision preserves motion semantics.

DINO-based self-supervision discovers partial motion structures, but different clusters exhibit substantial overlap in the physical space. This indicates that semantic visual features can capture high-level scene variations but are insufficient to fully distinguish controllable agent
motion from other visual changes. In contrast, Pixel-level self-supervision produces the most entangled clusters, where different latent clusters often correspond to similar
physical motions. This suggests that pixel-level objectives tend to encode appearance-related variations rather than navigation-specific dynamics.

Note that the presence of eight clusters alone does not indicate meaningful latent actions, since K-means always produces the predefined number of clusters. The key criterion is whether samples within the same cluster exhibit consistent physical effects. The results demonstrate that physical
supervision plays a critical role in grounding latent actions in controllable
navigation semantics.

\subsection{Cross-Dataset Consistency of Latent Action Semantics}
\label{sec:cross_dataset_consistency}

Beyond within-dataset motion alignment, we further evaluate whether latent action semantics remain consistent across heterogeneous environments and datasets. Unlike the previous analysis, which only examines motion structure within each dataset, this evaluation tests whether the learned latent action space provides a transferable navigation interface.

For each supervision paradigm, K-means clustering is first performed on latent representations from in-domain datasets using the original 32-dimensional latent space. The obtained cluster assignments are then fixed and directly
transferred to unseen datasets without additional clustering or adaptation. Specifically, we evaluate transfer from the in-domain datasets RECON, HuRoN, and SCAND to the out-of-domain datasets TartanDrive and Go Stanford.

To quantify semantic consistency, we discretize each continuous physical
motion transition $(\Delta x,\Delta y,\Delta\mathrm{yaw})$ into one of six
navigation categories, where $\Delta x$ and $\Delta y$ denote the two
planar displacement components and $\Delta\mathrm{yaw}$ denotes the
change in yaw angle. The category set is defined as
$\mathcal{C} =
\{\mathrm{F}, \mathrm{B}, \mathrm{FL}, \mathrm{FR}, \mathrm{BL}, \mathrm{BR}\}$,
corresponding to forward, backward, forward-left, forward-right,
backward-left, and backward-right, respectively.

Each latent cluster is labeled by its majority ground-truth motion category among in-domain samples, which determines predictions for samples in that cluster. We evaluate agreement with ground truth using Macro-6 Recall, the unweighted mean recall across all six categories:
\begin{equation}
\mathrm{Macro}\text{-}6\ \mathrm{Recall}
=
\frac{1}{6}
\sum_{c \in \mathcal{C}} R_c
=
\frac{1}{6}
\sum_{c \in \mathcal{C}}
\frac{\mathrm{TP}_c}
{\mathrm{TP}_c + \mathrm{FN}_c},
\label{eq:macro6_recall}
\end{equation}
where $R_c$ denotes the recall for category $c$,
$\mathrm{TP}_c$ is the number of samples whose ground-truth and predicted
categories are both $c$, and $\mathrm{FN}_c$ is the number of samples
whose ground-truth category is $c$ but whose predicted category differs
from $c$. This metric assigns equal weight to each motion category,
regardless of its frequency in the evaluation set.

\begin{figure}[H]
    \centering
     \includegraphics[width=0.8\textwidth]{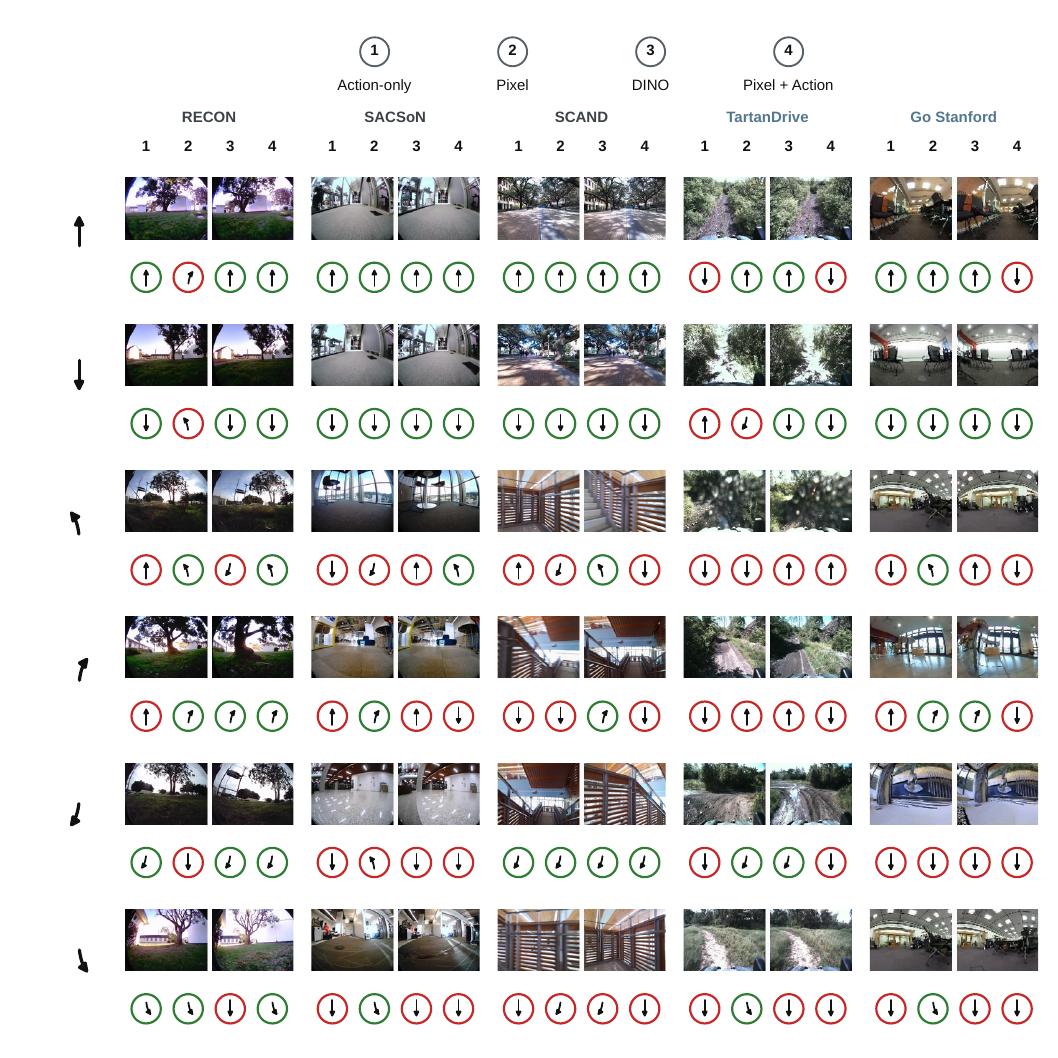}
     \caption{Visualization of the semantic separability of latent actions learned by four LAM paradigms across different navigation action categories. TartanDrive and Go Stanford are out-of-domain datasets.}
    \vspace{-12pt}
    \label{fig:lam_semantics}
\end{figure}

As shown in Figure~\ref{fig:lam_semantics}, Action-only supervision achieves strong motion-semantic consistency on
in-domain datasets, demonstrating that explicit motion supervision can align
latent actions with navigation behaviors. However, its performance decreases on unseen datasets, suggesting that motion semantics learned from
action-labeled data may still be affected by dataset-specific distributions.

Pixel-level self-supervision achieves weaker transfer performance, as visual
representations may capture scene appearance and observation variations
rather than consistent motion semantics. DINO-based self-supervision improves
semantic transferability but still exhibits ambiguity in fine-grained navigation motions.

Joint Action-Pixel supervision achieves the strongest cross-dataset consistency by combining scalable visual supervision with physically grounded motion annotations. This improves within-dataset motion alignment and preserves navigation semantics across heterogeneous environments, supporting UNLAs as a shared, transferable interface between visual experience and navigation dynamics.

\newpage

\section{Qualitative Results}
We present qualitative examples of OpenNWM's visual predictions, navigation trajectories, and LAM image reconstructions across diverse environments.

\begin{figure}[H]
    \centering
     \includegraphics[width=1\textwidth]{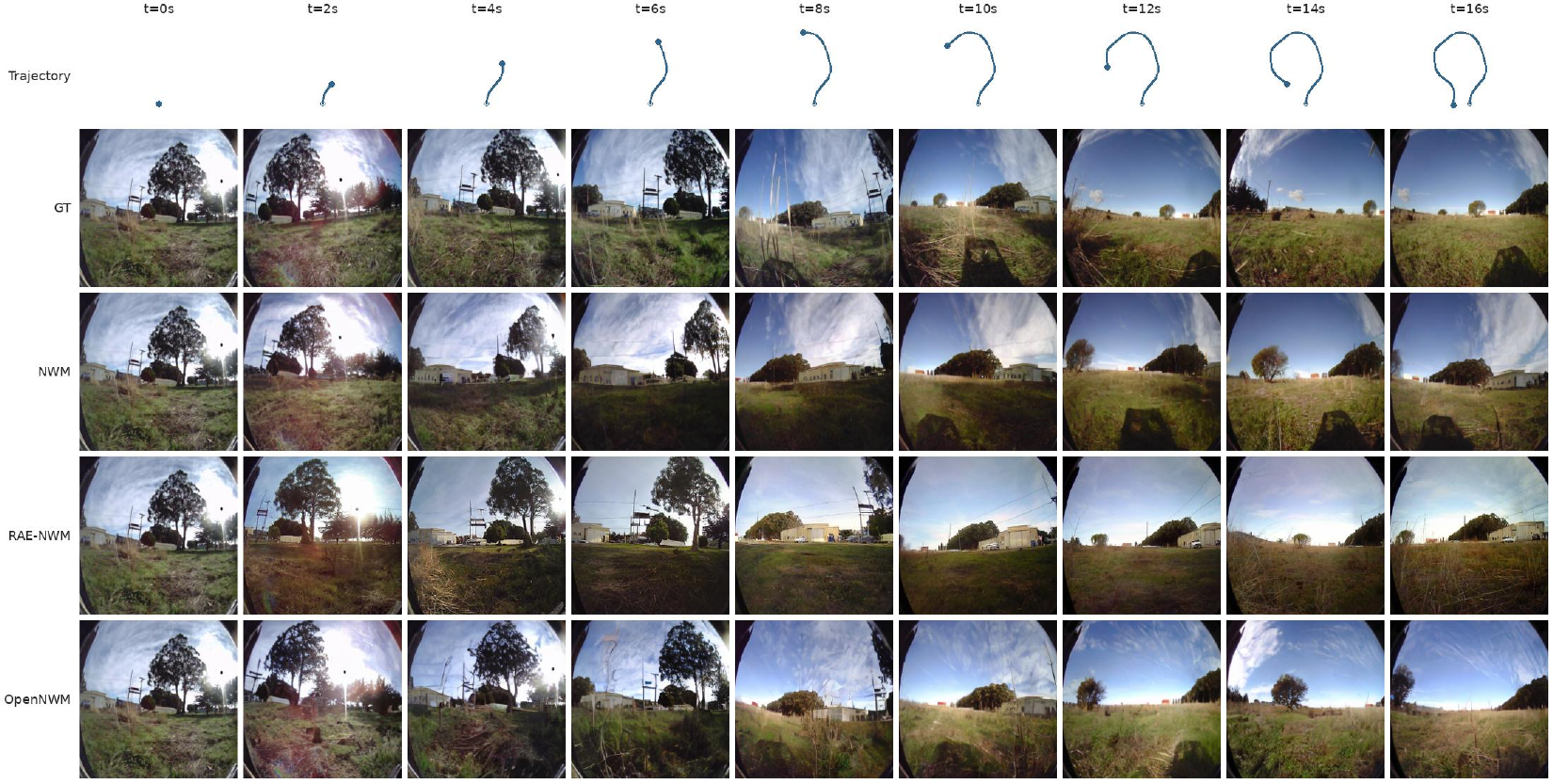}
    \caption{Video generation examples on RECON. NWM, RAE-NWM and OpenNWM are conditioned on a single first image, and a ground truth trajectory and
autoregressively predicts the next up to 16 seconds at 4 FPS. We plot the generated results from 0 to 16 seconds, every 2 second.}
    \vspace{-12pt}
\end{figure}

\begin{figure}[H]
    \centering
     \includegraphics[width=1\textwidth]{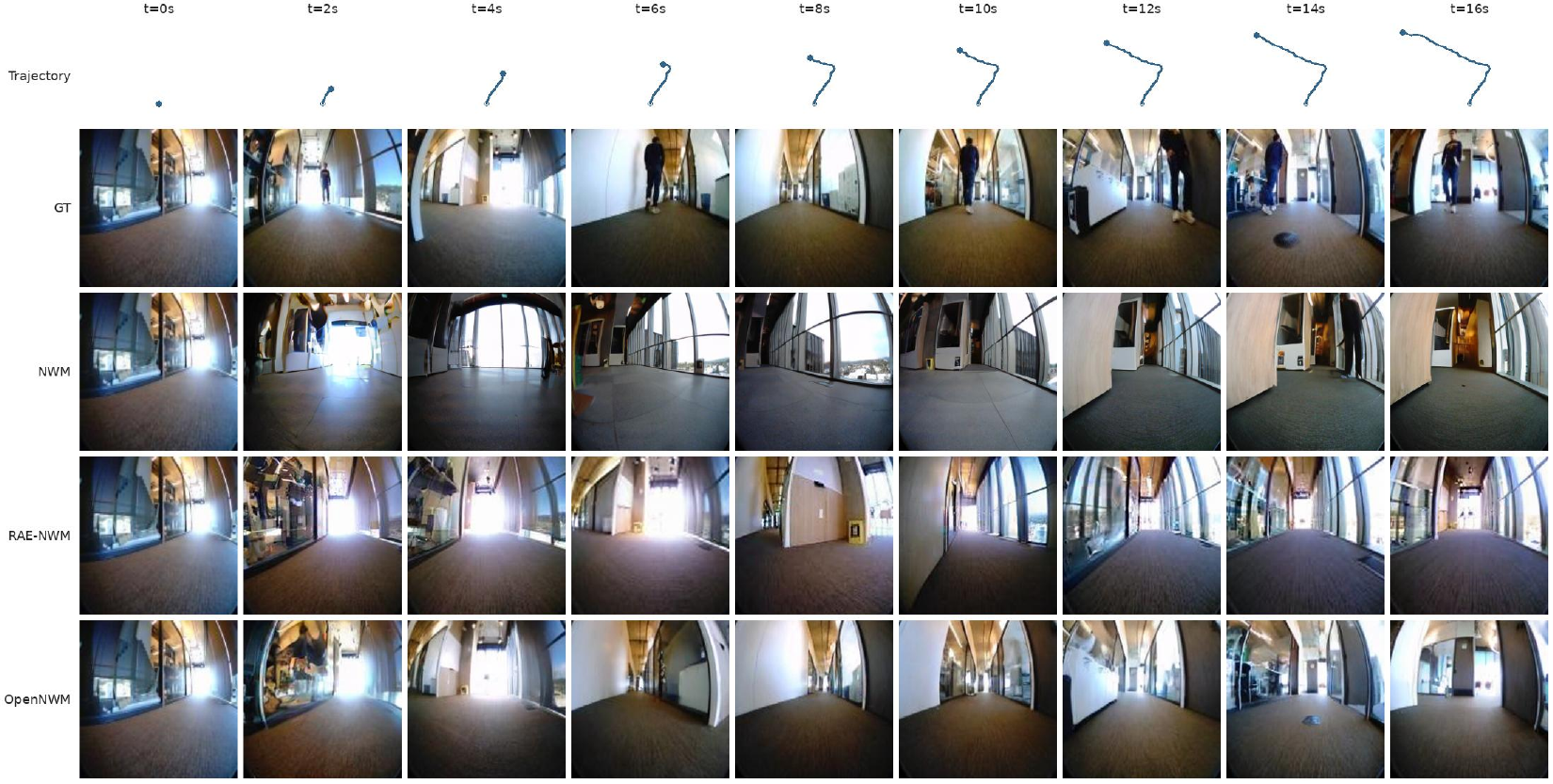}
    \caption{Video generation examples on HuRoN. NWM, RAE-NWM and OpenNWM are conditioned on a single first image, and a ground truth trajectory and
autoregressively predicts the next up to 16 seconds at 4 FPS. We plot the generated results from 0 to 16 seconds, every 2 second.}
    \vspace{-12pt}
\end{figure}

\begin{figure*}[t]
    \centering
     \includegraphics[width=1\textwidth]{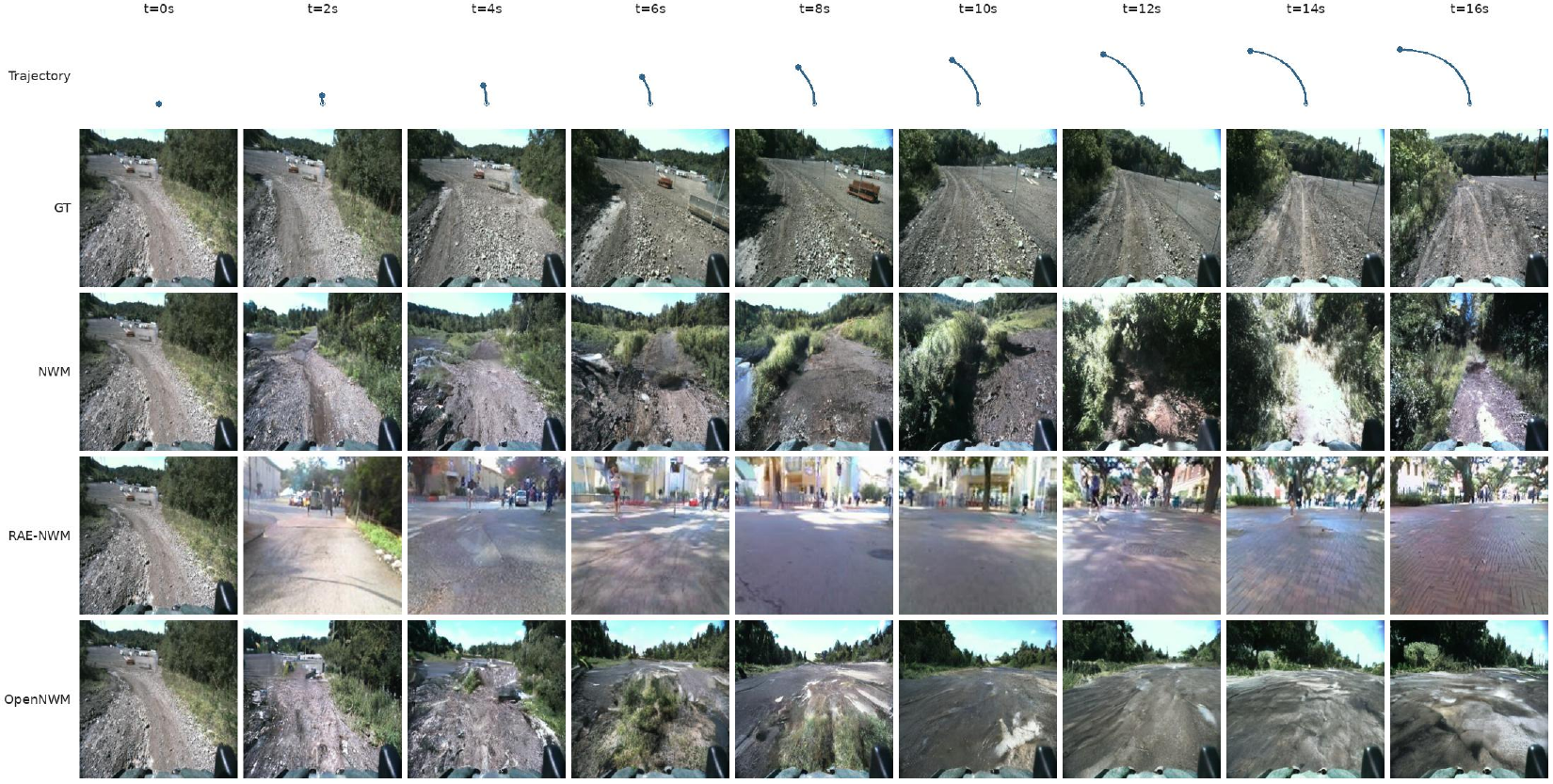}
    \caption{Video generation examples on TartanDrive. NWM, RAE-NWM and OpenNWM are conditioned on a single first image, and a ground truth trajectory and
autoregressively predicts the next up to 16 seconds at 4 FPS. We plot the generated results from 0 to 16 seconds, every 2 second.}
    \vspace{-12pt}
\end{figure*}

\begin{figure*}[t]
    \centering
     \includegraphics[width=1\textwidth]{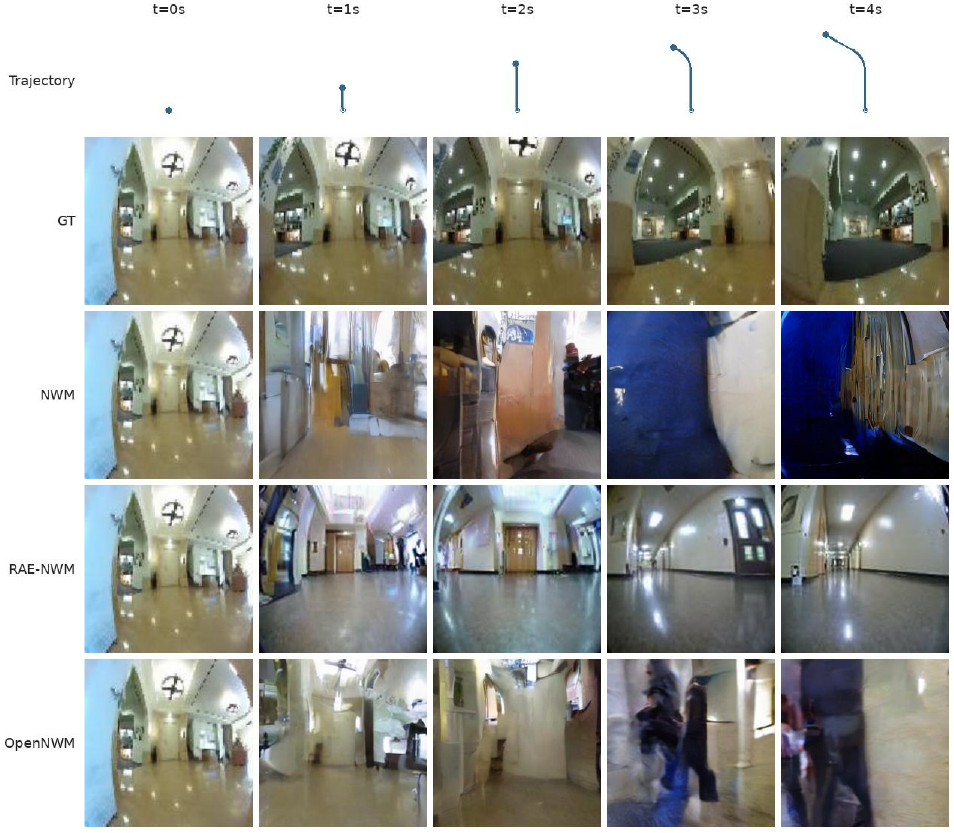}
    \caption{Video generation examples on Go Stanford (OOD). NWM, RAE-NWM and OpenNWM are conditioned on a single first image, and a ground truth trajectory and
autoregressively predicts the next up to 4 seconds at 4 FPS. We plot the generated results from 0 to 4 seconds, every 1 second.}
    \vspace{-12pt}
\end{figure*}


\begin{figure*}[t]
    \centering
    \includegraphics[width=1\textwidth]{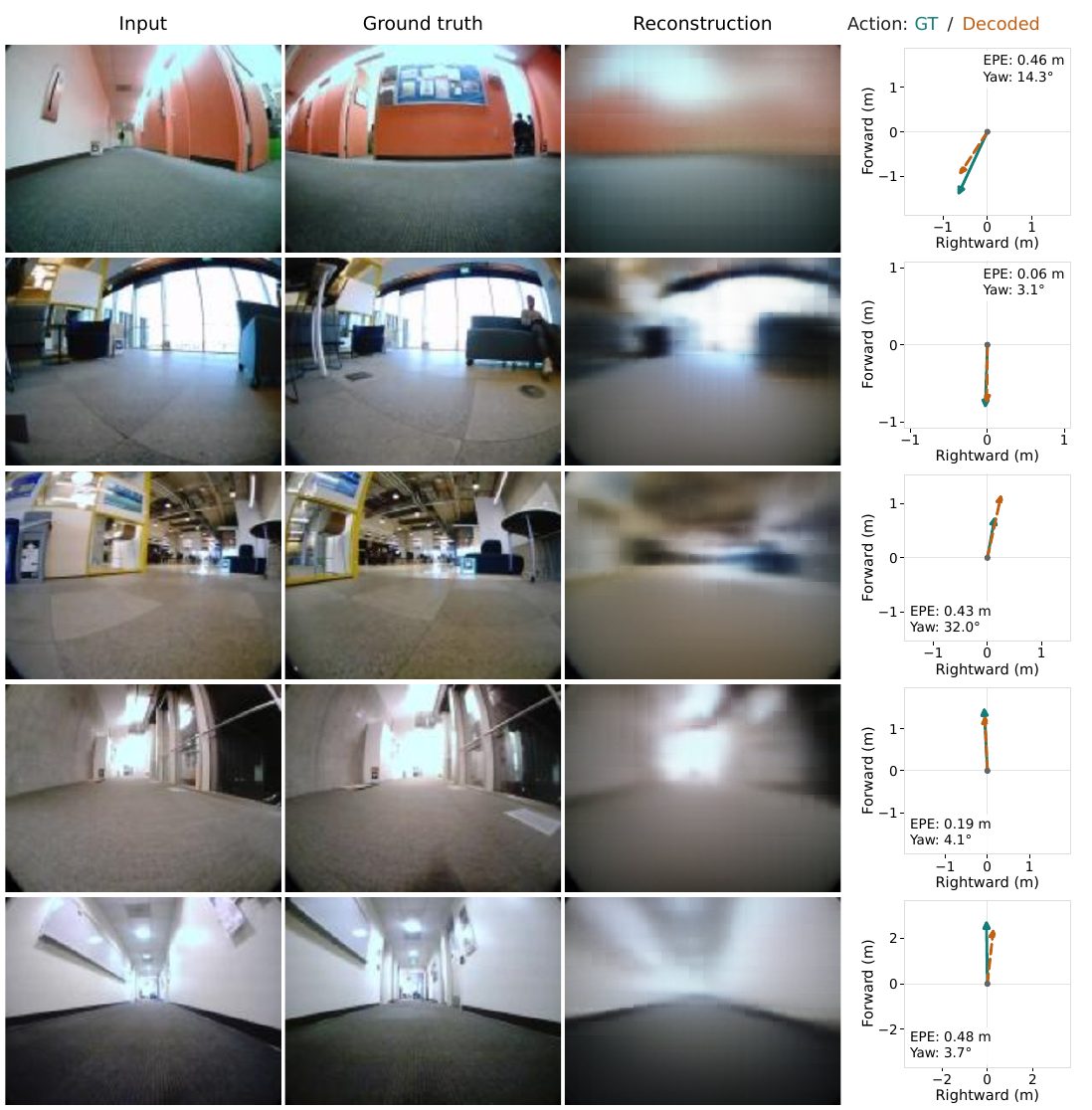}
    \caption{Qualitative results of our LAM on HuRoN. Each row shows the input image, ground-truth target image, reconstruction, and ground-truth versus decoded actions. Endpoint error (EPE, m) and yaw error ($^\circ$) are reported for each example.}
    \label{fig:vis-lam-huron}
    \vspace{-12pt}
\end{figure*}

\begin{figure*}[t]
    \centering
    \includegraphics[width=1\textwidth]{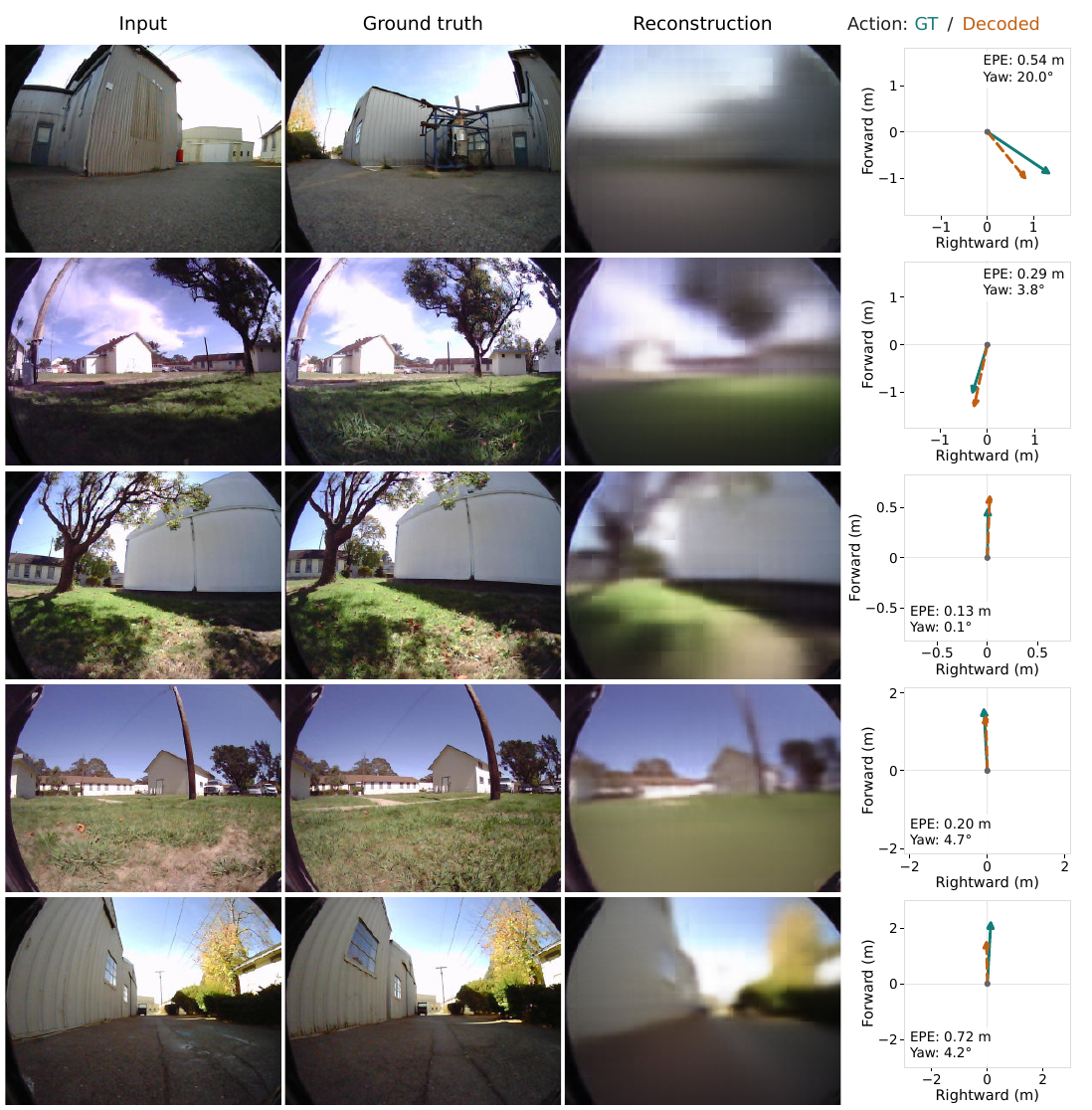}
    \caption{Qualitative results of our LAM on RECON. Each row shows the input image, ground-truth target image, reconstruction, and ground-truth versus decoded actions. Endpoint error (EPE, m) and yaw error ($^\circ$) are reported for each example.}
    \label{fig:vis-lam-recon}
    \vspace{-12pt}
\end{figure*}

\begin{figure*}[t]
    \centering
    \includegraphics[width=1\textwidth]{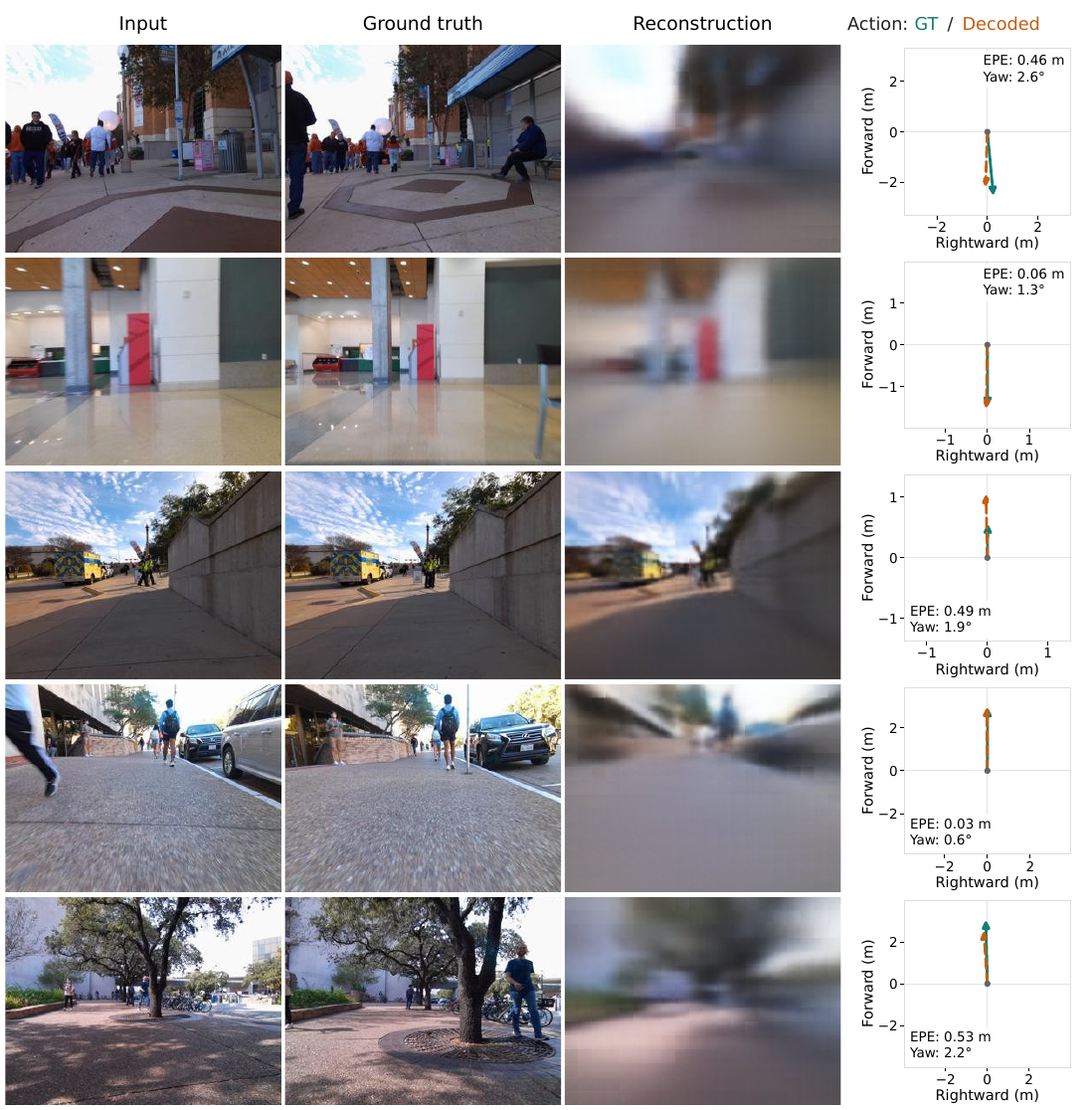}
    \caption{Qualitative results of our LAM on SCAND. Each row shows the input image, ground-truth target image, reconstruction, and ground-truth versus decoded actions. Endpoint error (EPE, m) and yaw error ($^\circ$) are reported for each example.}
    \label{fig:vis-lam-scand}
    \vspace{-12pt}
\end{figure*}

\begin{figure*}[t]
    \centering
    \includegraphics[width=1\textwidth]{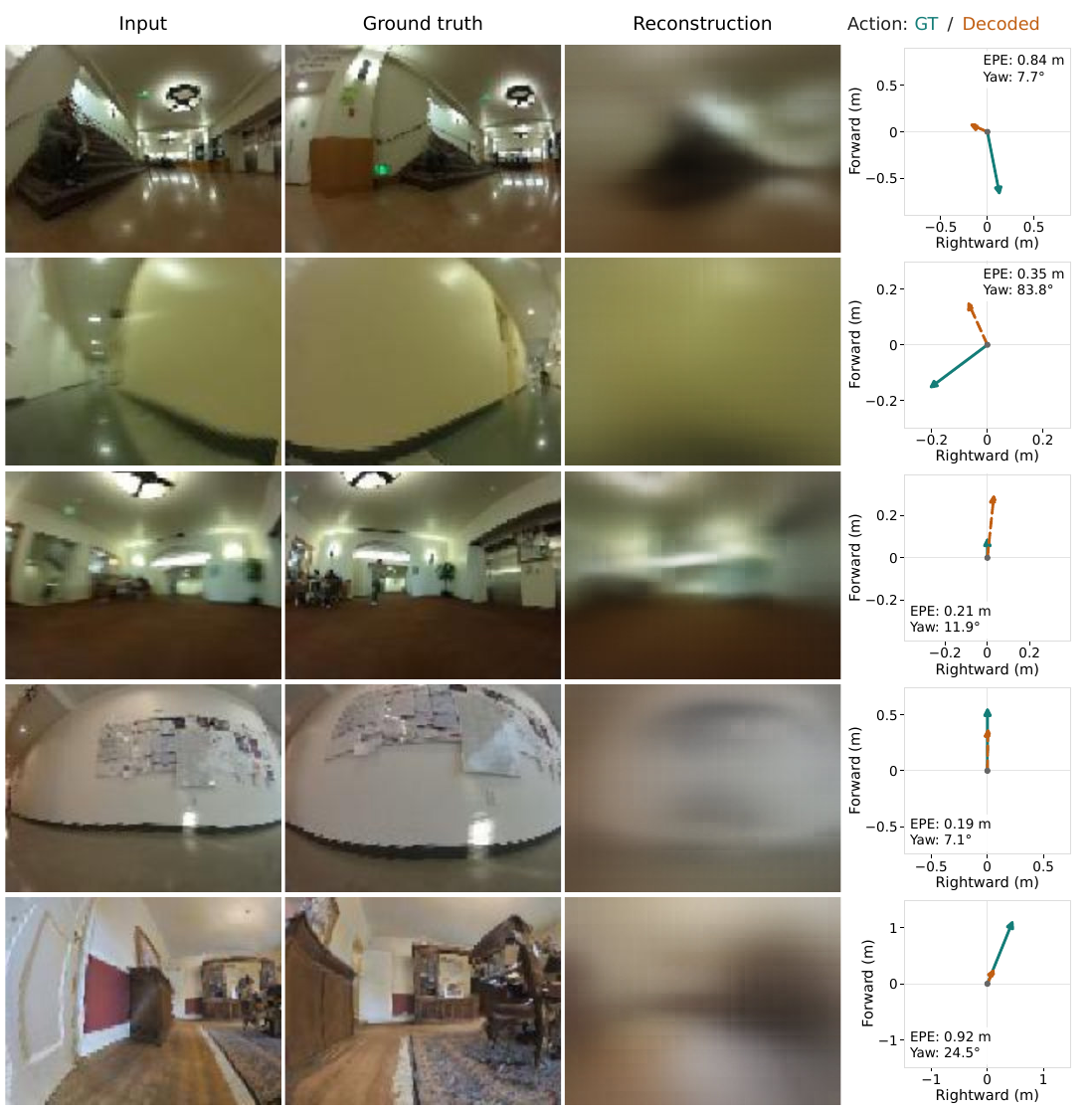}
    \caption{Qualitative results of our LAM on Go Stanford (OOD). Each row shows the input image, ground-truth target image, reconstruction, and ground-truth versus decoded actions. Endpoint error (EPE, m) and yaw error ($^\circ$) are reported for each example.}
    \label{fig:vis-lam-stanford}
    \vspace{-12pt}
\end{figure*}

\begin{figure*}[t]
    \centering
    \includegraphics[width=1\textwidth]{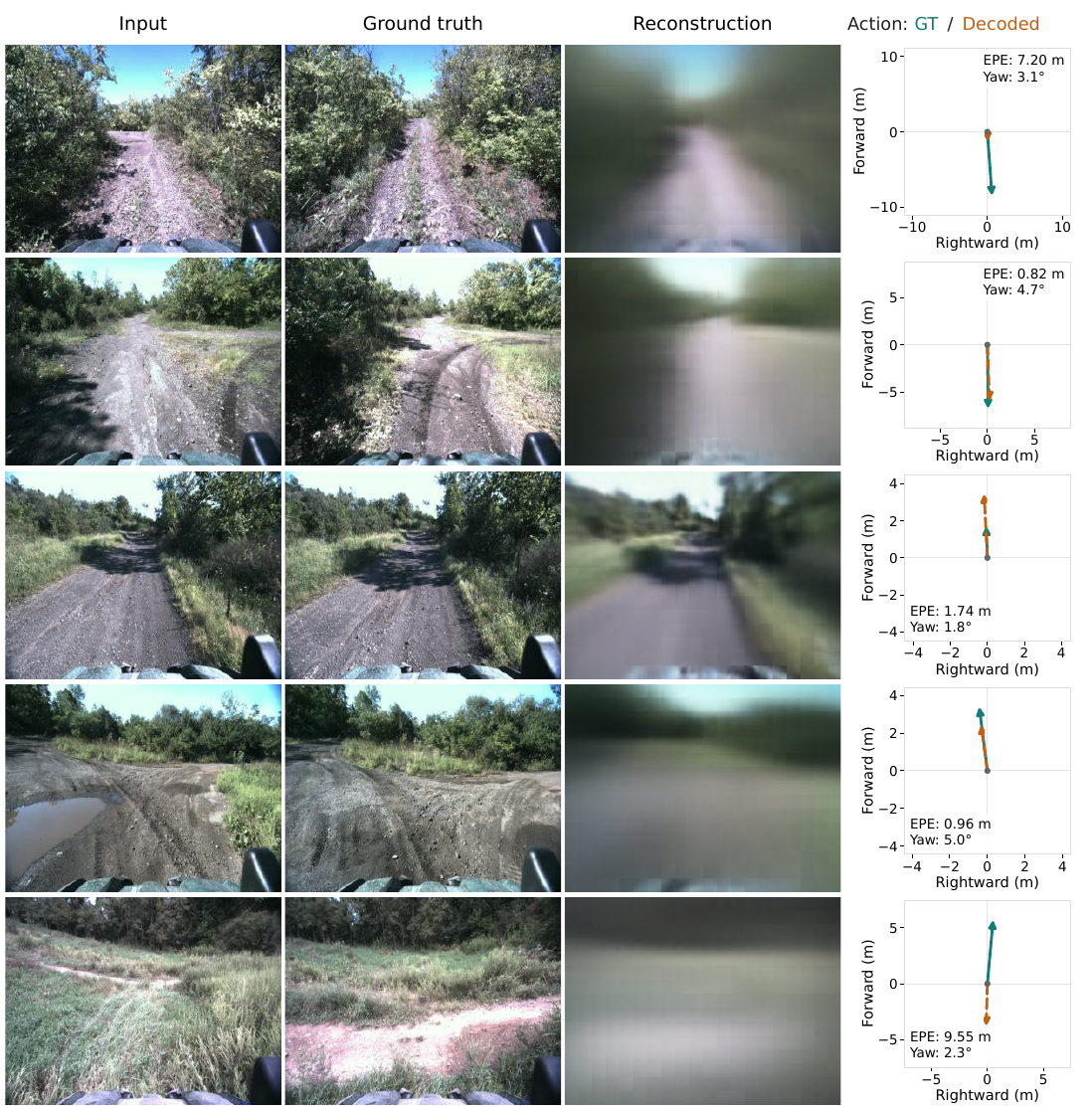}
    \caption{Qualitative results of our LAM on TartanDrive (OOD). Each row shows the input image, ground-truth target image, reconstruction, and ground-truth versus decoded actions. Endpoint error (EPE, m) and yaw error ($^\circ$) are reported for each example.}
    \label{fig:vis-lam-tartan}
    \vspace{-12pt}
\end{figure*}


\begin{figure*}[t]
    \centering
    \includegraphics[width=1\textwidth]{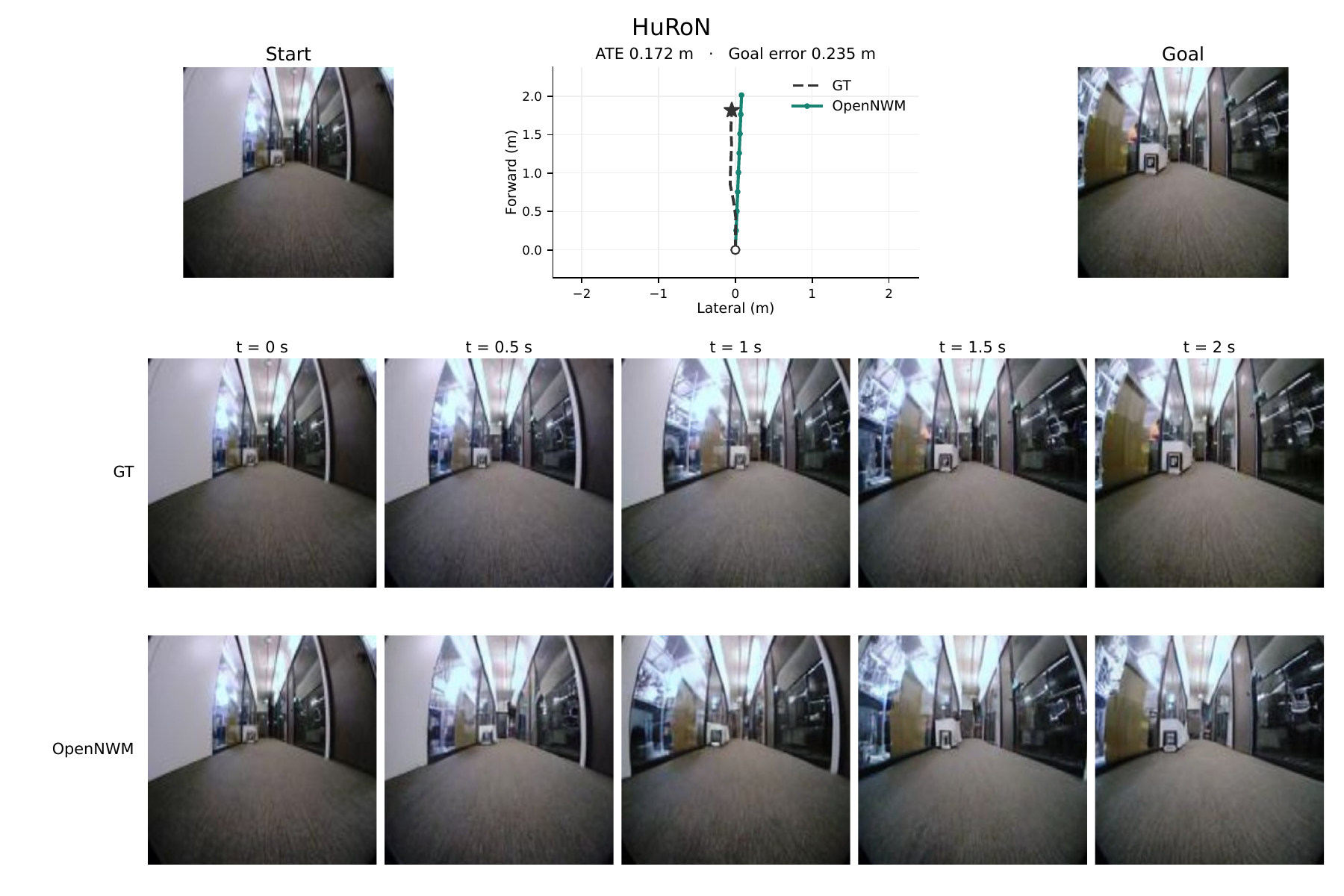}
    \caption{Qualitative navigation results of OpenNWM on HuRoN. The top row shows the start and goal images alongside ground-truth and predicted trajectories, with ATE and goal error reported in meters. The bottom rows compare ground-truth and predicted observations at 0.5-second intervals over a 2-second horizon.}
    \label{fig:vis-nav-huron-1}
    \vspace{-12pt}
\end{figure*}

\begin{figure*}[t]
    \centering
    \includegraphics[width=1\textwidth]{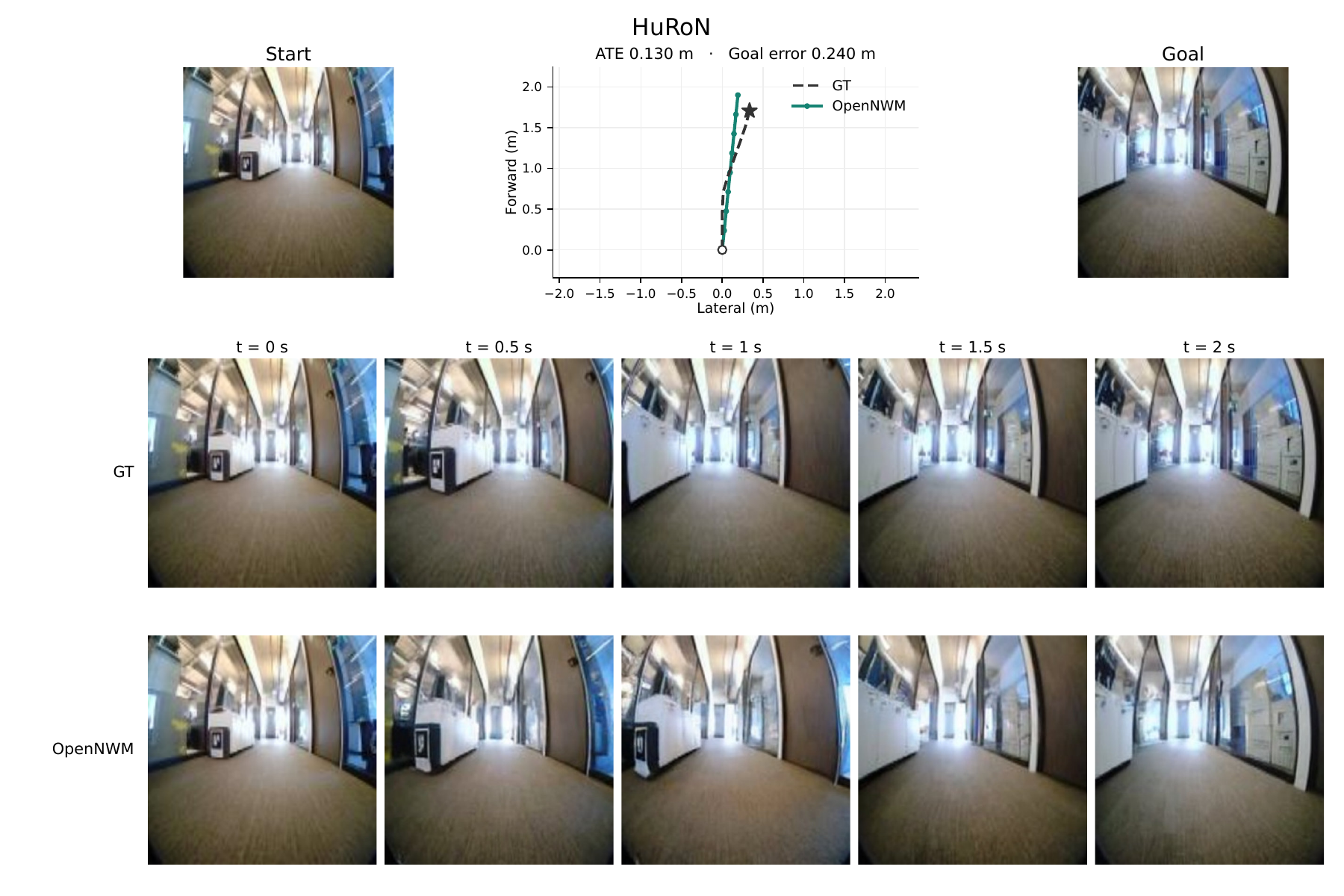}
    \caption{Qualitative navigation results of OpenNWM on HuRoN. The top row shows the start and goal images alongside ground-truth and predicted trajectories, with ATE and goal error reported in meters. The bottom rows compare ground-truth and predicted observations at 0.5-second intervals over a 2-second horizon.}
    \label{fig:vis-nav-huron-2}
    \vspace{-12pt}
\end{figure*}

\begin{figure*}[t]
    \centering
    \includegraphics[width=1\textwidth]{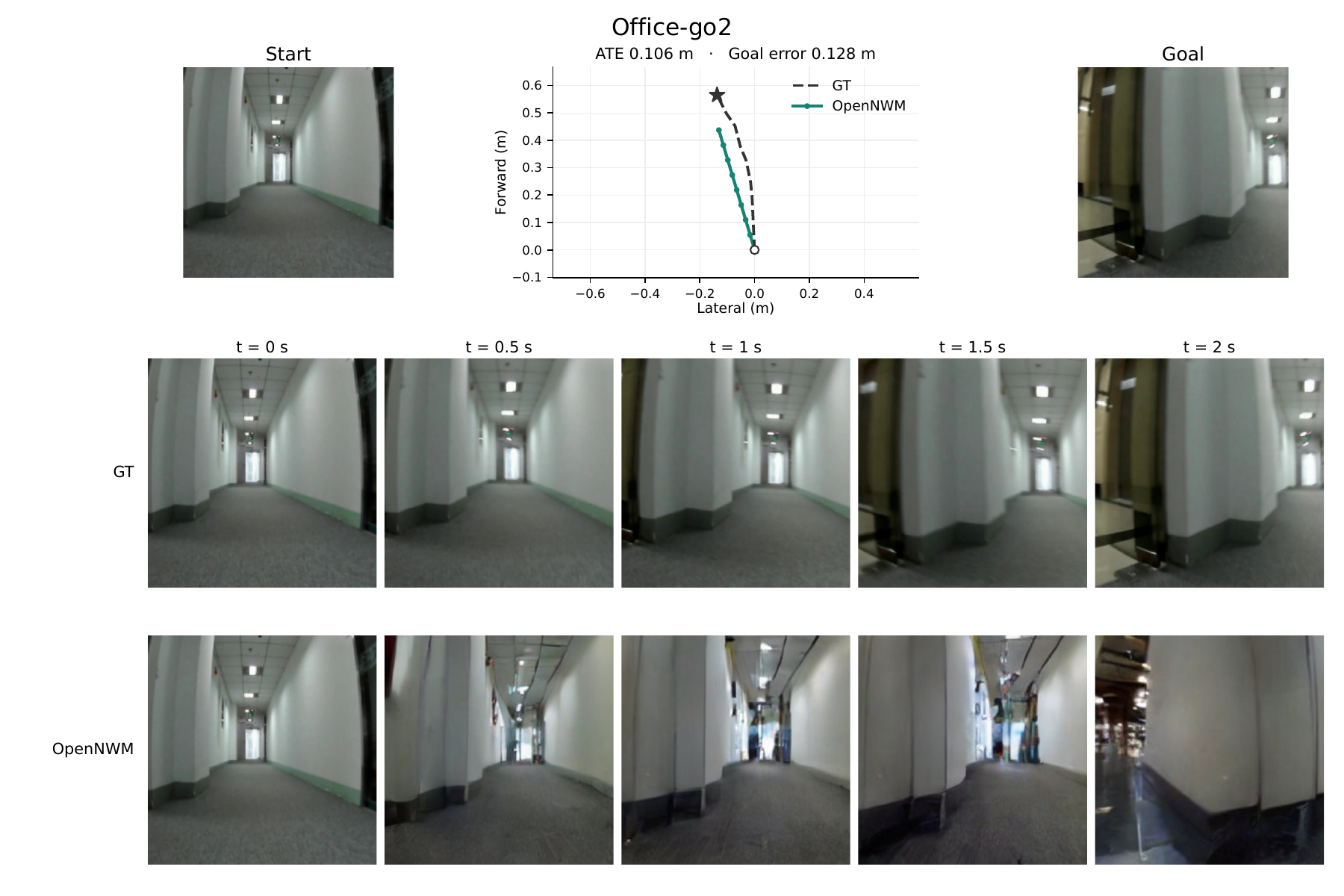}
    \caption{Qualitative navigation results of OpenNWM on Office-go2 (OOD). The top row shows the start and goal images alongside ground-truth and predicted trajectories, with ATE and goal error reported in meters. The bottom rows compare ground-truth and predicted observations at 0.5-second intervals over a 2-second horizon.}
    \label{fig:vis-nav-office-go2-1}
    \vspace{-12pt}
\end{figure*}

\begin{figure*}[t]
    \centering
    \includegraphics[width=1\textwidth]{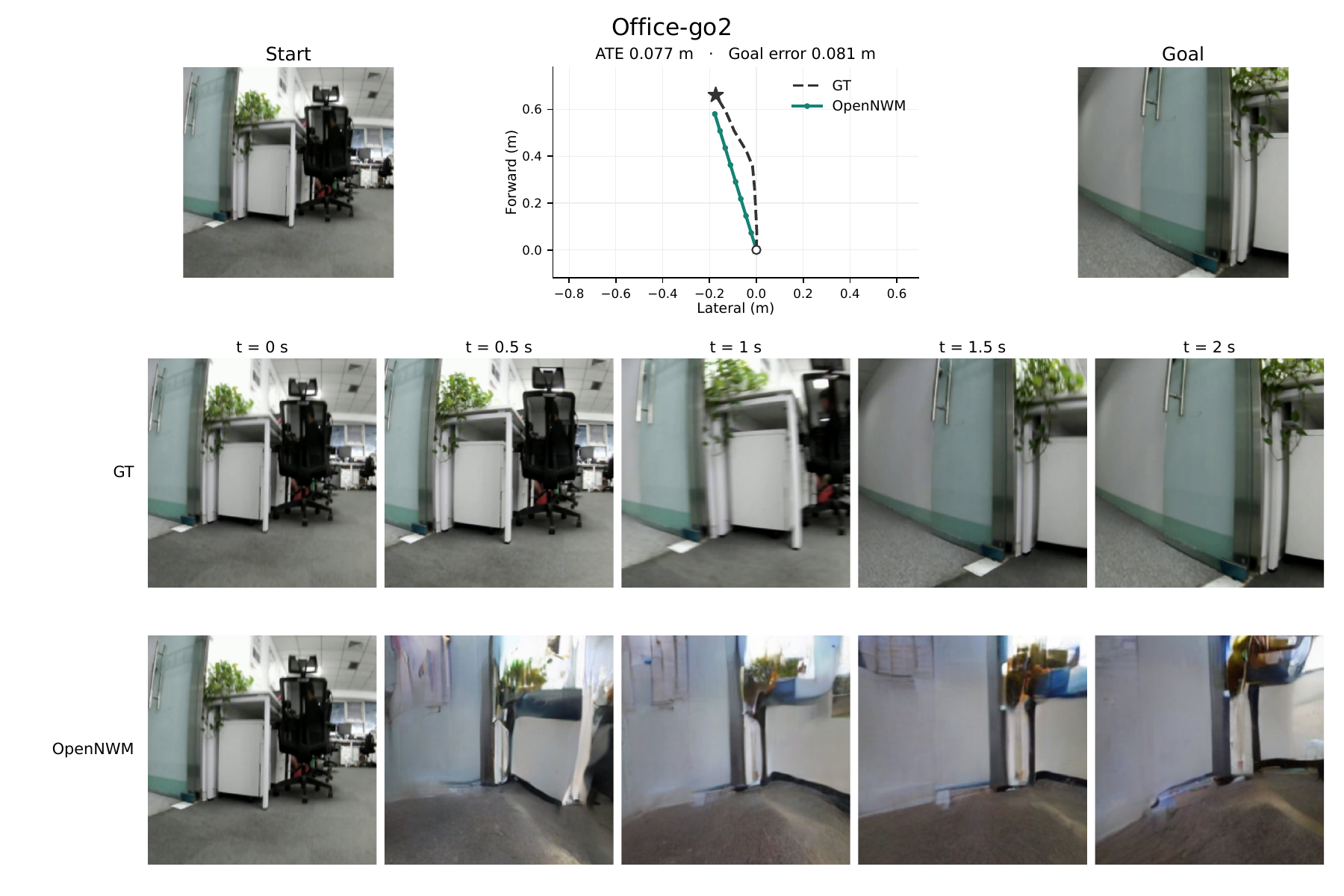}
    \caption{Qualitative navigation results of OpenNWM on Office-go2 (OOD). The top row shows the start and goal images alongside ground-truth and predicted trajectories, with ATE and goal error reported in meters. The bottom rows compare ground-truth and predicted observations at 0.5-second intervals over a 2-second horizon.}
    \label{fig:vis-nav-office-go2-2}
    \vspace{-12pt}
\end{figure*}

\begin{figure*}[t]
    \centering
    \includegraphics[width=1\textwidth]{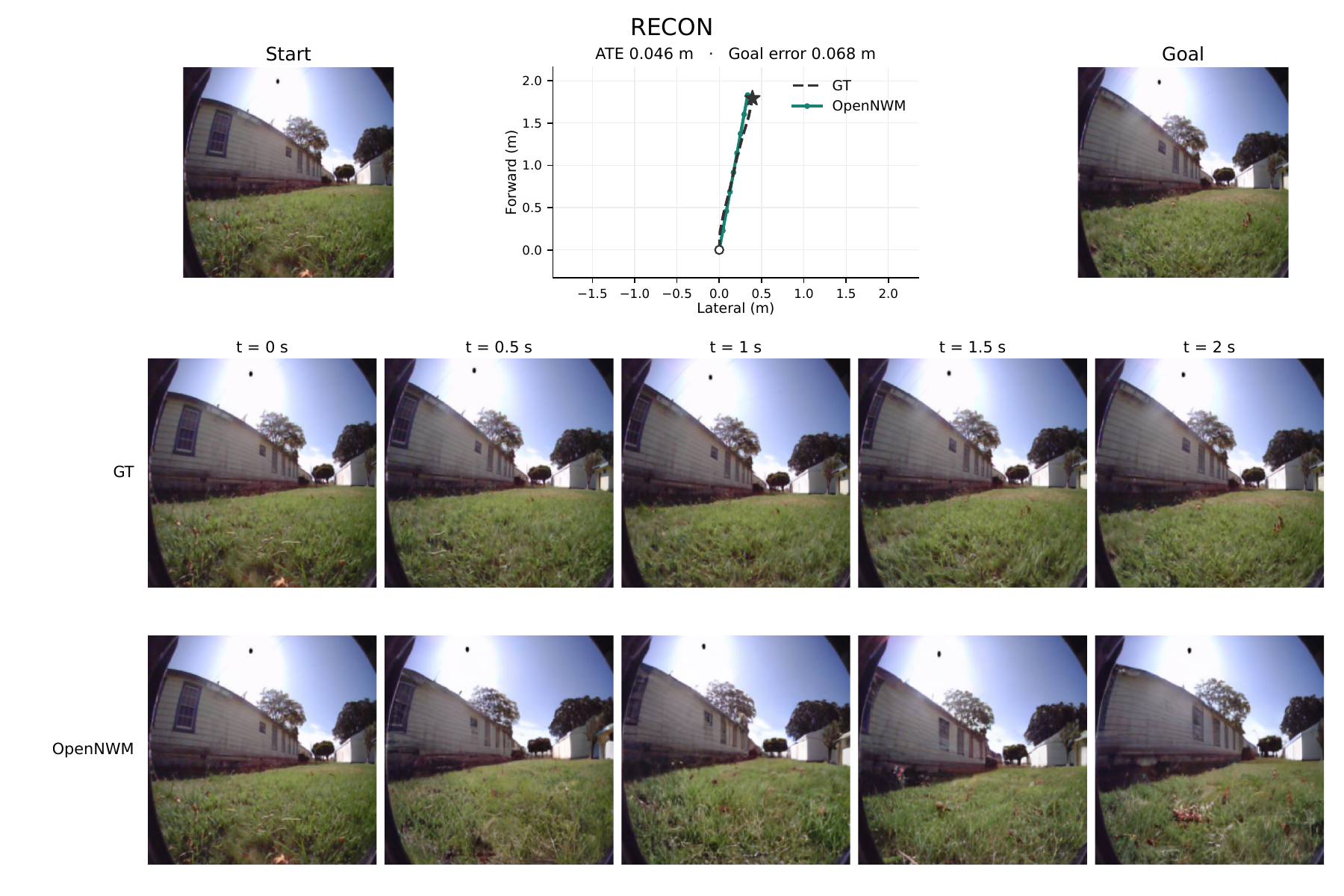}
    \caption{Qualitative navigation results of OpenNWM on RECON. The top row shows the start and goal images alongside ground-truth and predicted trajectories, with ATE and goal error reported in meters. The bottom rows compare ground-truth and predicted observations at 0.5-second intervals over a 2-second horizon.}
    \label{fig:vis-nav-recon-1}
    \vspace{-12pt}
\end{figure*}

\begin{figure*}[t]
    \centering
    \includegraphics[width=1\textwidth]{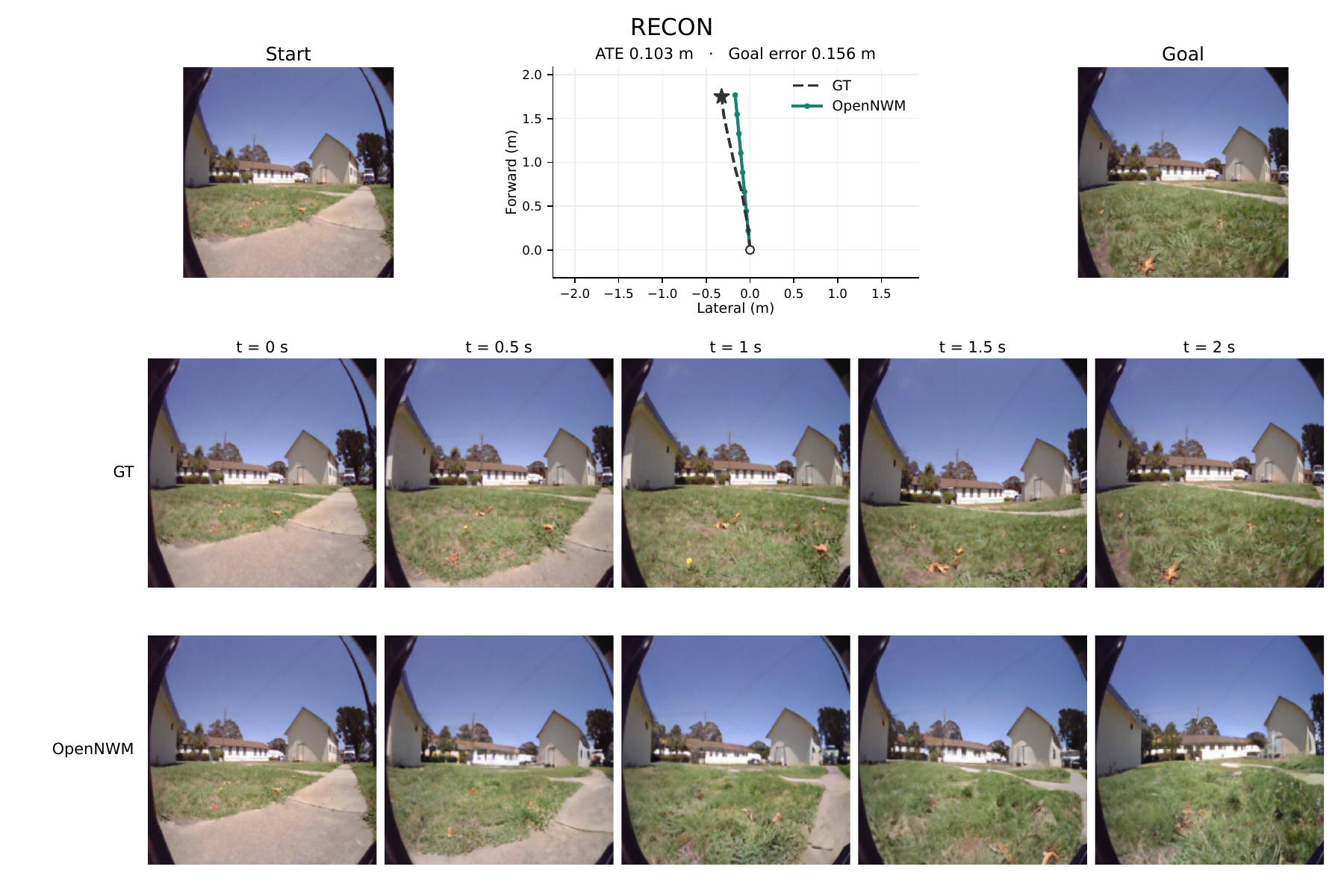}
    \caption{Qualitative navigation results of OpenNWM on RECON. The top row shows the start and goal images alongside ground-truth and predicted trajectories, with ATE and goal error reported in meters. The bottom rows compare ground-truth and predicted observations at 0.5-second intervals over a 2-second horizon.}
    \label{fig:vis-nav-recon-2}
    \vspace{-12pt}
\end{figure*}

\begin{figure*}[t]
    \centering
    \includegraphics[width=1\textwidth]{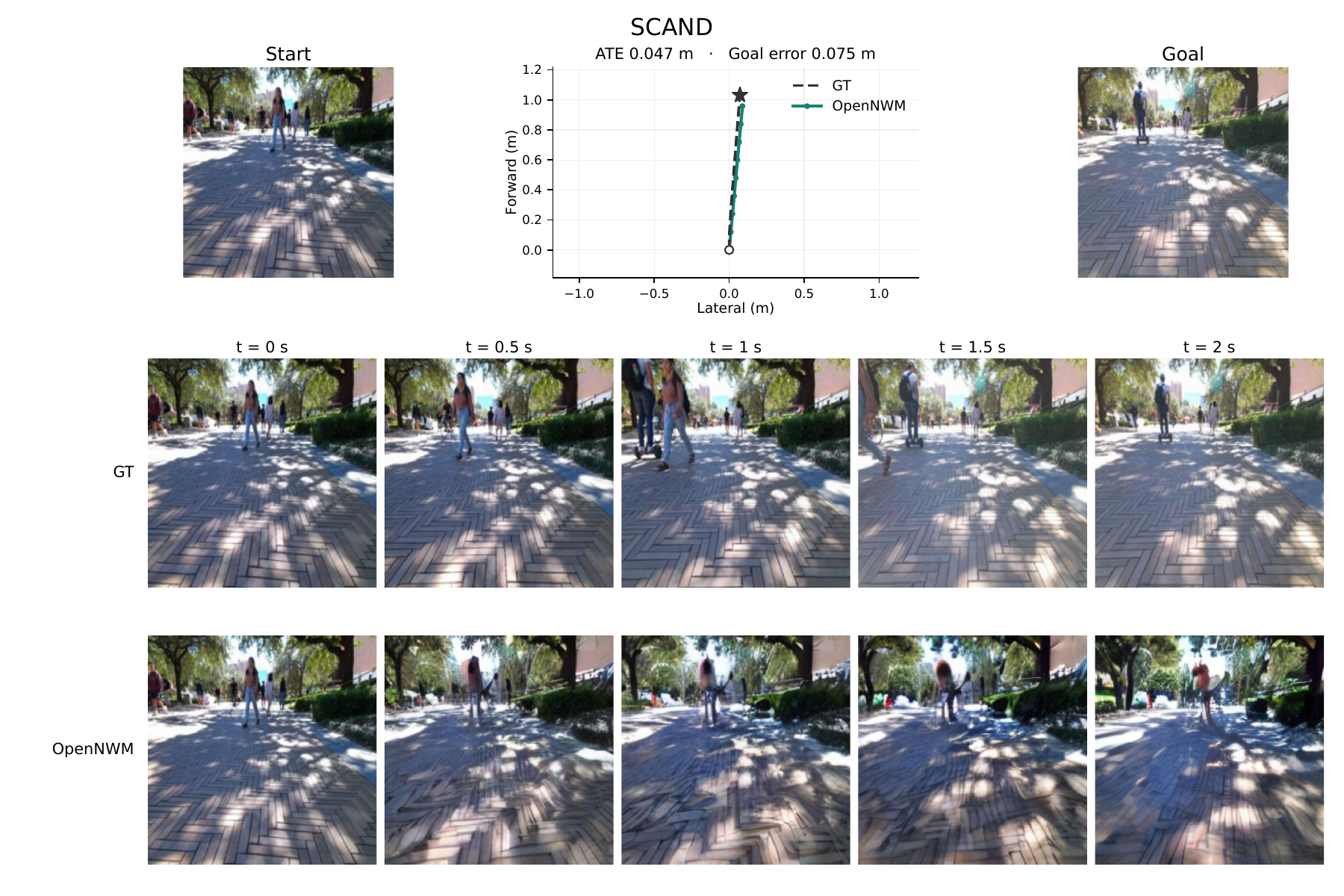}
    \caption{Qualitative navigation results of OpenNWM on SCAND. The top row shows the start and goal images alongside ground-truth and predicted trajectories, with ATE and goal error reported in meters. The bottom rows compare ground-truth and predicted observations at 0.5-second intervals over a 2-second horizon.}
    \label{fig:vis-nav-scand-1}
    \vspace{-12pt}
\end{figure*}

\begin{figure*}[t]
    \centering
    \includegraphics[width=1\textwidth]{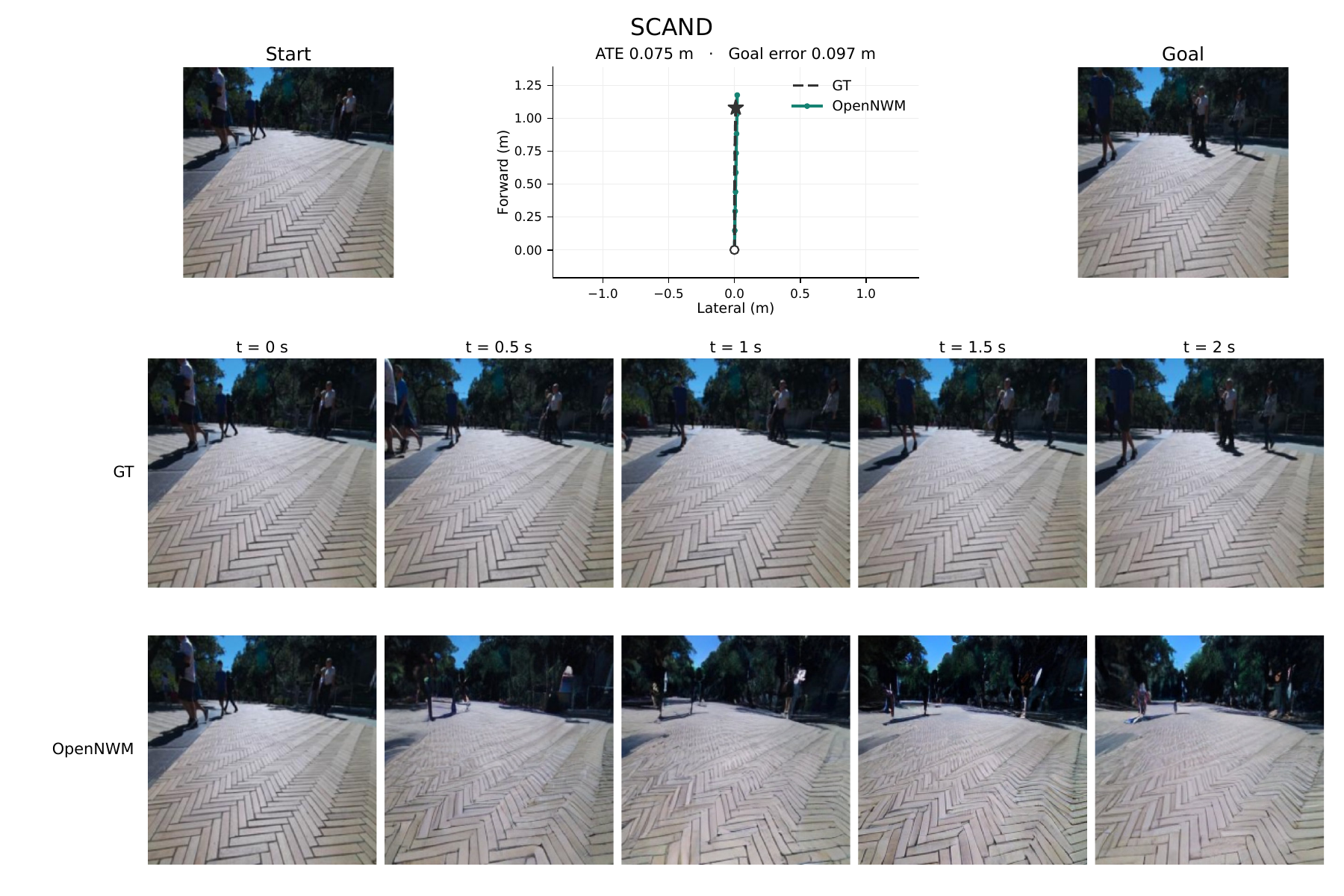}
    \caption{Qualitative navigation results of OpenNWM on SCAND. The top row shows the start and goal images alongside ground-truth and predicted trajectories, with ATE and goal error reported in meters. The bottom rows compare ground-truth and predicted observations at 0.5-second intervals over a 2-second horizon.}
    \label{fig:vis-nav-scand-2}
    \vspace{-12pt}
\end{figure*}

\begin{figure*}[t]
    \centering
    \includegraphics[width=1\textwidth]{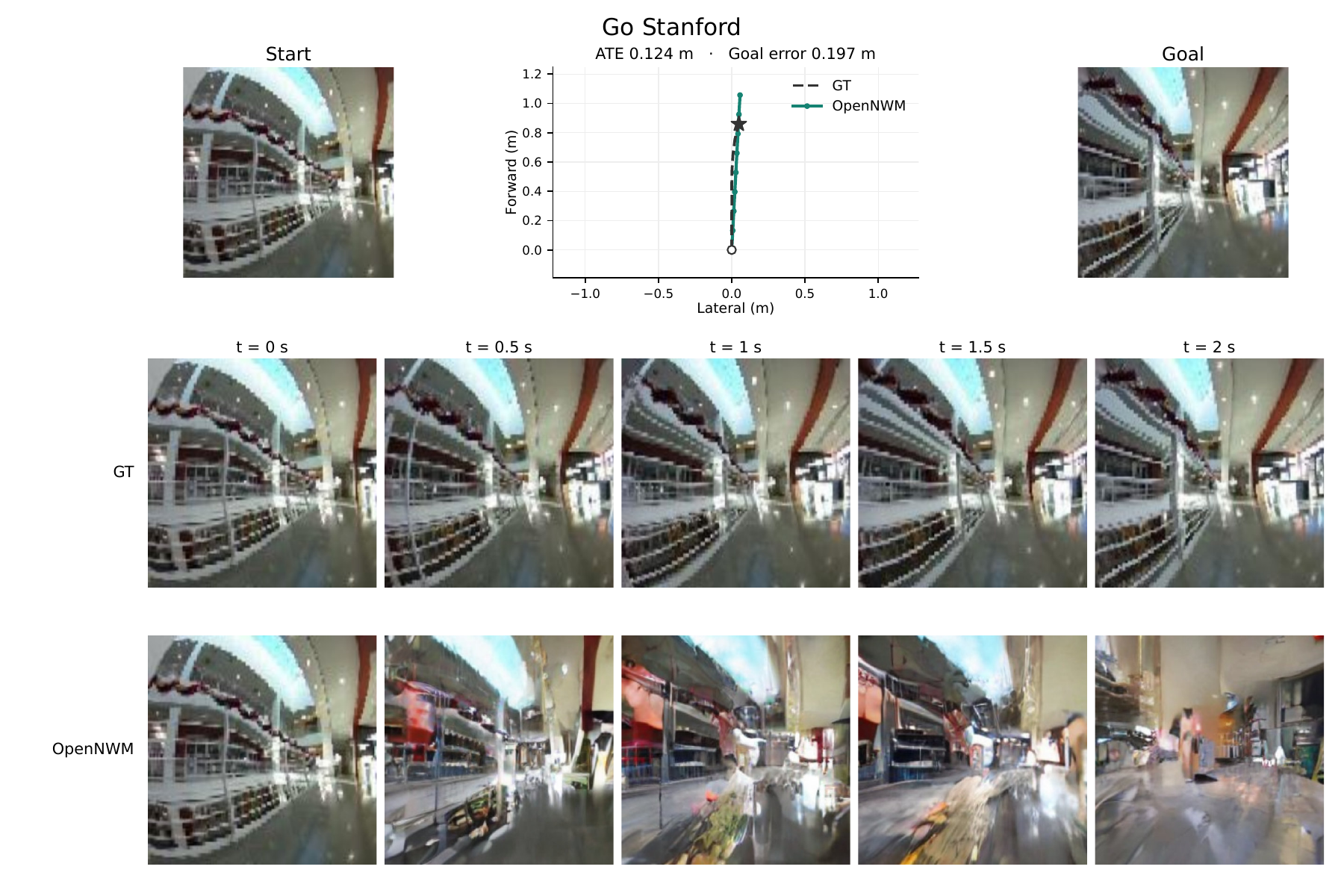}
    \caption{Qualitative navigation results of OpenNWM on Go Stanford (OOD). The top row shows the start and goal images alongside ground-truth and predicted trajectories, with ATE and goal error reported in meters. The bottom rows compare ground-truth and predicted observations at 0.5-second intervals over a 2-second horizon.}
    \label{fig:vis-nav-stanford-1}
    \vspace{-12pt}
\end{figure*}

\begin{figure*}[t]
    \centering
    \includegraphics[width=1\textwidth]{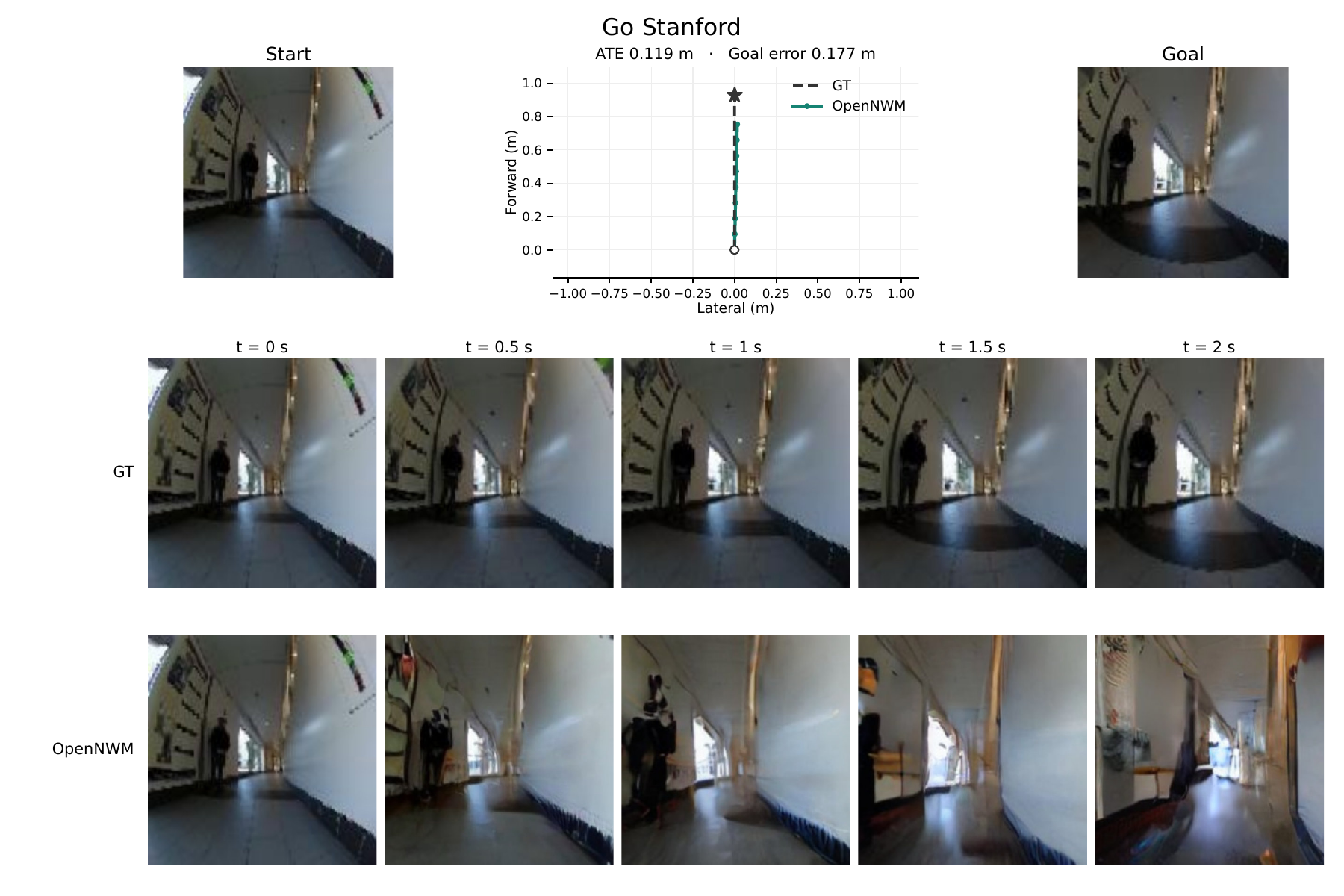}
    \caption{Qualitative navigation results of OpenNWM on Go Stanford (OOD). The top row shows the start and goal images alongside ground-truth and predicted trajectories, with ATE and goal error reported in meters. The bottom rows compare ground-truth and predicted observations at 0.5-second intervals over a 2-second horizon.}
    \label{fig:vis-nav-stanford-2}
    \vspace{-12pt}
\end{figure*}

\begin{figure*}[t]
    \centering
    \includegraphics[width=1\textwidth]{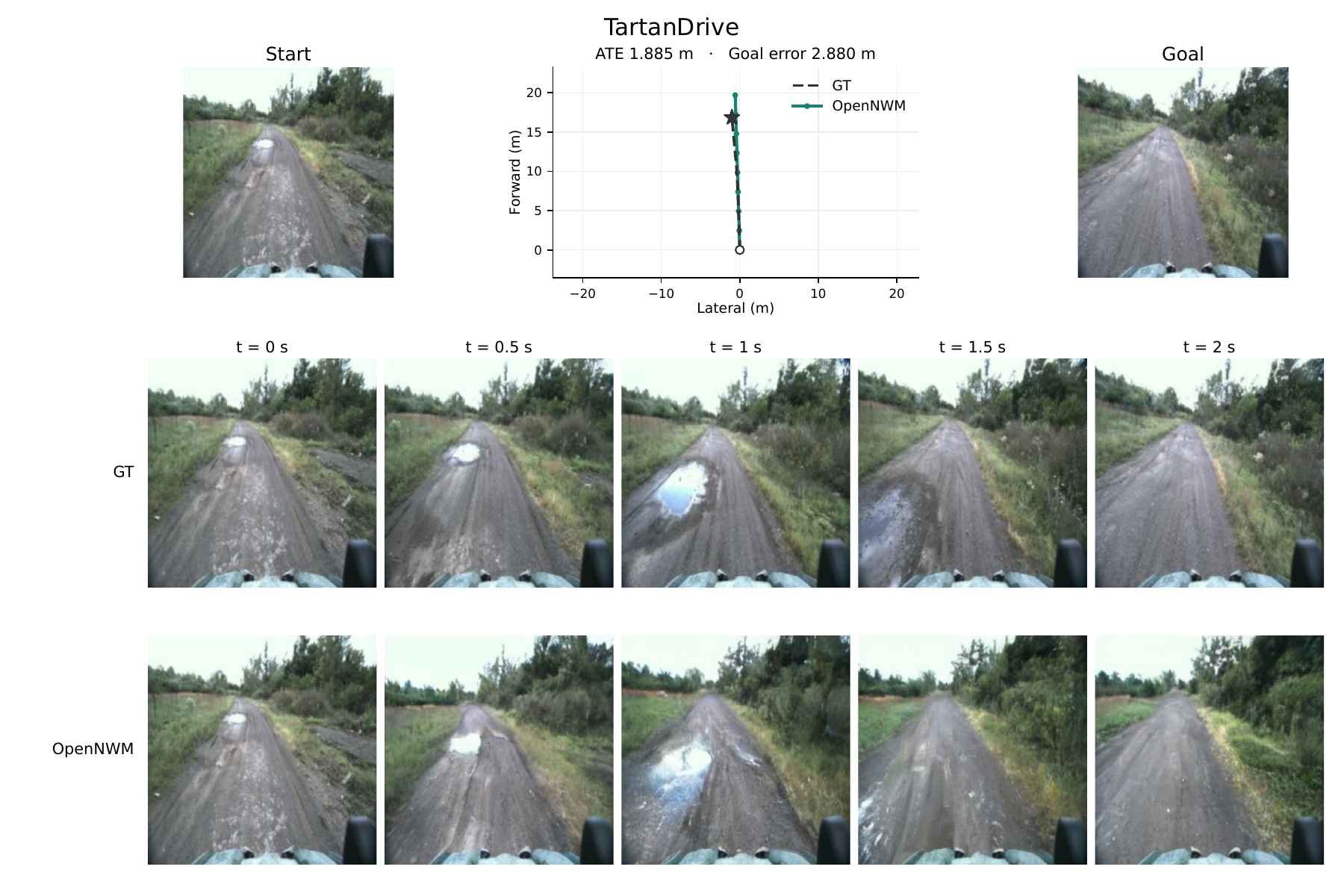}
    \caption{Qualitative navigation results of OpenNWM on TartanDrive. The top row shows the start and goal images alongside ground-truth and predicted trajectories, with ATE and goal error reported in meters. The bottom rows compare ground-truth and predicted observations at 0.5-second intervals over a 2-second horizon.}
    \label{fig:vis-nav-tartan-1}
    \vspace{-12pt}
\end{figure*}

\begin{figure*}[t]
    \centering
    \includegraphics[width=1\textwidth]{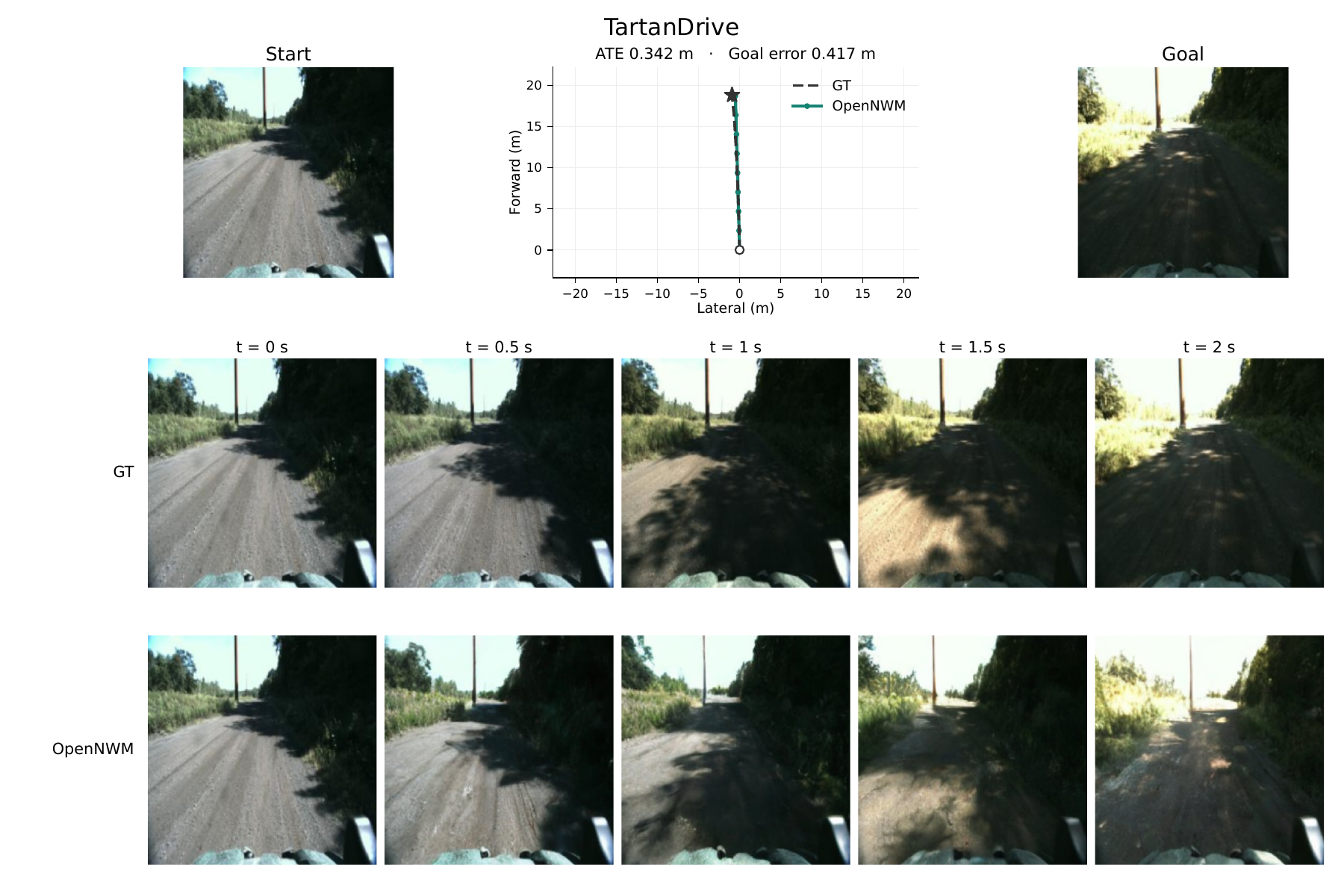}
    \caption{Qualitative navigation results of OpenNWM on TartanDrive. The top row shows the start and goal images alongside ground-truth and predicted trajectories, with ATE and goal error reported in meters. The bottom rows compare ground-truth and predicted observations at 0.5-second intervals over a 2-second horizon.}
    \label{fig:vis-nav-tartan-2}
    \vspace{-12pt}
\end{figure*}

%% file: tables/paras.tex
\begin{table}[t]
\centering
\small
\caption{\textbf{Training and inference configurations of OpenNWM.}
Batch sizes are global, and image resolutions are reported as
height $\times$ width.}
\label{tab:training_details}
\small
\setlength{\tabcolsep}{24pt}
\renewcommand{\arraystretch}{1.02}
\resizebox{0.9\linewidth}{!}{%
\begin{tabular}{@{}lll@{}}
\toprule
\textbf{Module} & \textbf{Configuration} & \textbf{Value} \\
\midrule

\multirow{17}{*}{\makecell[l]{Latent Action\\Model}}
& Architecture & Pixel+Action spatiotemporal Transformer \\
& Model size & ${\sim}709$M \\
& Encoder / decoder blocks & 24 / 24 \\
& Hidden width / attention heads & 1,024 / 16 \\
& Latent-action dimension & 32 \\
& Input frames & 2 \\
& Input resolution / patch size & $240\times320$ / $16\times16$ \\
& Temporal offsets & Integers in $[-8,8]$ \\
& Action-labeled batch fraction & 25\% \\
& RGB / action / KL loss weights & $1$ / $0.171244$ / $10^{-6}$ \\
& Training steps & 60K \\
& Global batch size & 160 \\
& Optimizer & AdamW \\
& Peak learning rate / weight decay & $2.5\times10^{-5}$ / $0.01$ \\
& LR warm-up / schedule length & 12.8K / 100K steps \\
& LR decay / minimum LR ratio & Cosine / $0.1$ \\
& Precision / training hardware & BF16 / $8\times$ NVIDIA L20 \\
\midrule

\multirow{15}{*}{\makecell[l]{Navigation\\World Model}}
& Backbone / model size & CDiT-B/2 / ${\sim}197$M \\
& Image tokenizer & Frozen SD-VAE \\
& Input resolution & $224\times224$ \\
& Context frames / targets per observation & 4 / 4 \\
& Target frame offsets & Integers in $[-64,64]$ \\
& Latent-action conditioning range & $|\Delta t|\leq8$ \\
& Pretraining steps & 60K \\
& Pretraining learning rate & $2\times10^{-4}$ \\
& Physical-action adapter warm-up & 3K steps \\
& Joint post-training steps & 100K \\
& Post-training LR: adapter / backbone & $10^{-4}$ / $10^{-4}$ \\
& Global observation / prediction batch & 384 / 1,536 \\
& Optimizer / weight decay & AdamW / $0.01$ \\
& Gradient clipping & 2.0 \\
& Precision / training hardware & BF16 / $8\times$ NVIDIA L20 \\
\midrule

\multirow{9}{*}{\makecell[l]{Inference and\\Planning}}
& Diffusion sampler / sampling steps & DDPM / 250 \\
& Planning action representation & Physical motion $(\Delta x,\Delta y,\Delta\psi)$ \\
& Planning horizon & $2\,\mathrm{s}$ \\
& Action interval / sequence length & $0.25\,\mathrm{s}$ / 8 \\
& CEM candidates & 80 \\
& CEM elites & 5 \\
& CEM updates & 1 \\
& Stochastic evaluations per candidate & 3 \\
& Planning cost & LPIPS-Alex \\
\bottomrule
\end{tabular}%
}
\end{table}

%% file: tables/appendix_id.tex

%

\begin{table}[t]
\centering
\caption{\textbf{In-distribution (ID) direct visual prediction
at the nominal 4\,s horizon.}
DS: DreamSim.
Best and second-best results are bold and underlined, respectively.}
\label{tab:id_direct_4s}

\small
\setlength{\tabcolsep}{5pt}
\renewcommand{\arraystretch}{1}

\resizebox{0.96\linewidth}{!}{%
\begin{tabular}{@{}l*{12}{c}@{}}
\toprule
\multirow{2}{*}{\textbf{Method}}
& \multicolumn{3}{c}{\textbf{RECON}}
& \multicolumn{3}{c}{\textbf{SCAND}}
& \multicolumn{3}{c}{\textbf{HuRoN}}
& \multicolumn{3}{c}{\textbf{TartanDrive}} \\
\cmidrule(lr){2-4}
\cmidrule(lr){5-7}
\cmidrule(lr){8-10}
\cmidrule(lr){11-13}
& LPIPS$\downarrow$ & DS$\downarrow$ & PSNR$\uparrow$
& LPIPS$\downarrow$ & DS$\downarrow$ & PSNR$\uparrow$
& LPIPS$\downarrow$ & DS$\downarrow$ & PSNR$\uparrow$
& LPIPS$\downarrow$ & DS$\downarrow$ & PSNR$\uparrow$ \\
\midrule

NWM
& \underline{0.323} & \underline{0.120} & 15.487
& 0.406 & 0.300 & 12.778
& 0.463 & 0.275 & 13.377
& 0.427 & 0.223 & \underline{13.602} \\

NWM XL (with Ego4D)
& 0.359 & 0.140 & 14.850
& \underline{0.394} & \underline{0.260} & 13.068
& 0.501 & 0.295 & 12.604
& \underline{0.424} & \underline{0.213} & 13.569 \\

RAE-NWM
& 0.353 & 0.134 & \underline{15.491}
& 0.413 & \textbf{0.190} & \textbf{14.419}
& \textbf{0.339} & \textbf{0.171} & \textbf{15.079}
& 0.662 & 0.519 & 10.840 \\

\midrule

\textbf{OpenNWM}
& \textbf{0.320} & \textbf{0.116} & \textbf{15.614}
& \textbf{0.385} & 0.271 & \underline{13.247}
& \underline{0.362} & \underline{0.193} & \underline{14.329}
& \textbf{0.405} & \textbf{0.203} & \textbf{14.005} \\

\bottomrule
\end{tabular}%
}
\end{table}

%% file: tables/appendix_navigation.tex

\begin{table}[t]
\centering
\caption{Navigation performance on RECON, SCAND, and TartanDrive.
Best and second-best available results are bold and underlined, respectively.}
\label{tab:main_navigation}

\small
\setlength{\tabcolsep}{15pt}
\renewcommand{\arraystretch}{1}

\resizebox{0.75\linewidth}{!}{%
\begin{tabular}{@{}l*{6}{c}@{}}
\toprule
\multirow{2}{*}{\textbf{Method}}
& \multicolumn{2}{c}{\textbf{RECON}}
& \multicolumn{2}{c}{\textbf{SCAND}}
& \multicolumn{2}{c}{\textbf{TartanDrive}} \\
\cmidrule(lr){2-3}
\cmidrule(lr){4-5}
\cmidrule(lr){6-7}
& ATE$\downarrow$
& RPE$\downarrow$
& ATE$\downarrow$
& RPE$\downarrow$
& ATE$\downarrow$
& RPE$\downarrow$ \\
\midrule

GNM
& 1.87 & 0.73
& 2.12 & 0.61
& 6.65 & 1.62 \\

NoMaD
& 1.95 & 0.53
& 2.24 & 0.49
& 6.32 & 1.31 \\

NWM
& \underline{1.13} & \textbf{0.35}
& \underline{1.01} & \underline{0.28}
& \underline{5.72} & \underline{1.19} \\

RAE-NWM
&1.36& \underline{0.37} 
& 1.14 & \underline{0.28}
& 5.99 & 1.25 \\

Garrido et al.
& 1.40 & 0.40
& -- & --
& -- & -- \\

CompACT
& 1.33 & 0.39
& 1.36 & 0.34
& -- & -- \\

\midrule

\textbf{OpenNWM (Ours)}
& \textbf{1.12} & \textbf{0.35}
& \textbf{0.98} & \textbf{0.27}
& \textbf{5.53} & \textbf{1.16} \\

\bottomrule
\end{tabular}%
}

\end{table}